\documentclass[essd, manuscript]{copernicus}

\usepackage{subfig}
\usepackage{enumitem}
\usepackage{booktabs}
\usepackage{multirow}
\usepackage{array}
\usepackage{graphicx}
\usepackage[acronym]{glossaries}
\usepackage{verbatim}

\usepackage{caption}
\usepackage{afterpage} 
\usepackage{float}

\usepackage[utf8]{inputenc} 
\usepackage[T1]{fontenc}    
\usepackage{hyperref}       
\usepackage{url}            
\usepackage{amsfonts}       
\usepackage{nicefrac}       
\usepackage{microtype}      
\usepackage{xcolor}         
\usepackage{graphicx}
\usepackage{wrapfig}
\usepackage{float}
\usepackage{multicol}
\usepackage{placeins}
\usepackage{longtable}
\usepackage{siunitx}
\colorlet{blue}{black}

\makeglossaries

\graphicspath{ {figs/} }

\begin{document}
\nolinenumbers

\title{GlobalGeoTree: A Multi-Granular Vision-Language Dataset for Global Tree Species Classification}


\Author[1,6]{Yang}{Mu}
\Author[1,6]{Zhitong}{Xiong}
\Author[1]{Yi}{Wang}
\Author[2]{Muhammad}{Shahzad}
\Author[3]{Franz}{Essl}
\Author[4]{Holger}{Kreft}
\Author[5]{Mark}{van Kleunen}
\Author[1,6][xiaoxiang.zhu@tum.de]{Xiao Xiang}{Zhu}

\affil[1]{Technical University of Munich, Munich, Germany}
\affil[2]{University of Reading, Reading, UK}
\affil[3]{University of Vienna, Vienna, Austria}
\affil[4]{University of Göttingen, Göttingen, Germany}
\affil[5]{University of Konstanz, Konstanz, Germany}
\affil[6]{Munich Center for Machine Learning}

\correspondence{Xiao Xiang Zhu (xiaoxiang.zhu@tum.de)}





\runningtitle{GlobalGeoTree}

\runningauthor{X.X.Z.} 

\received{}
\pubdiscuss{} 
\revised{}
\accepted{}
\published{}


\firstpage{1}

\maketitle

\begin{abstract}
Global tree species mapping using remote sensing data is vital for biodiversity monitoring, forest management, and ecological research. However, progress in this field has been constrained by the scarcity of large-scale, labeled datasets. To address this, we introduce GlobalGeoTree, a comprehensive global dataset for tree species classification. GlobalGeoTree comprises 6.3 million geolocated tree occurrences, spanning 275 families, 2,734 genera, and 21,001 species across the hierarchical taxonomic levels. Each sample is paired with Sentinel-2 image time series and 27 auxiliary environmental variables, encompassing bioclimatic, geographic, and soil data. The dataset is partitioned into \textit{GlobalGeoTree-6M}, a large subset for model pretraining, and curated evaluation subsets, primarily \textit{GlobalGeoTree-10kEval}, a benchmark for zero-shot and few-shot classification. To demonstrate the utility of the dataset, we introduce a baseline model, GeoTreeCLIP, which leverages paired remote sensing data and taxonomic text labels within a vision-language framework pretrained on \textit{GlobalGeoTree-6M}. Experimental results show that GeoTreeCLIP achieves substantial improvements in zero- and few-shot classification on \textit{GlobalGeoTree-10kEval} over existing advanced models. By making the dataset, models, and code publicly available, we aim to establish a benchmark to advance tree species classification and foster innovation in biodiversity research and ecological applications. The code is publicly available at \url{https://github.com/MUYang99/GlobalGeoTree}, and the GlobalGeoTree dataset is available at \url{https://huggingface.co/datasets/yann111/GlobalGeoTree} \citep{doi2025globalgeotree}.   
\end{abstract}


\section{Introduction}
Forests cover approximately 31\% of the global land surface \citep{hansen2013high} and provide essential ecosystem services, including carbon sequestration \citep{jenkins2003national}, biodiversity conservation \citep{lindenmayer2006general}, and climate regulation \citep{bonan2008forests}. Accurate and large-scale mapping of tree species plays an increasingly vital role in addressing pressing environmental challenges \citep{mu2025national}, including effective biodiversity monitoring \citep{felton2020tree}, informed forest management practices \citep{franklin2001remote}, and comprehensive ecological research aimed at understanding the complex impacts of climate change \citep{hamann2006potential}.

\begin{figure}
\centering
\includegraphics[width=\textwidth]{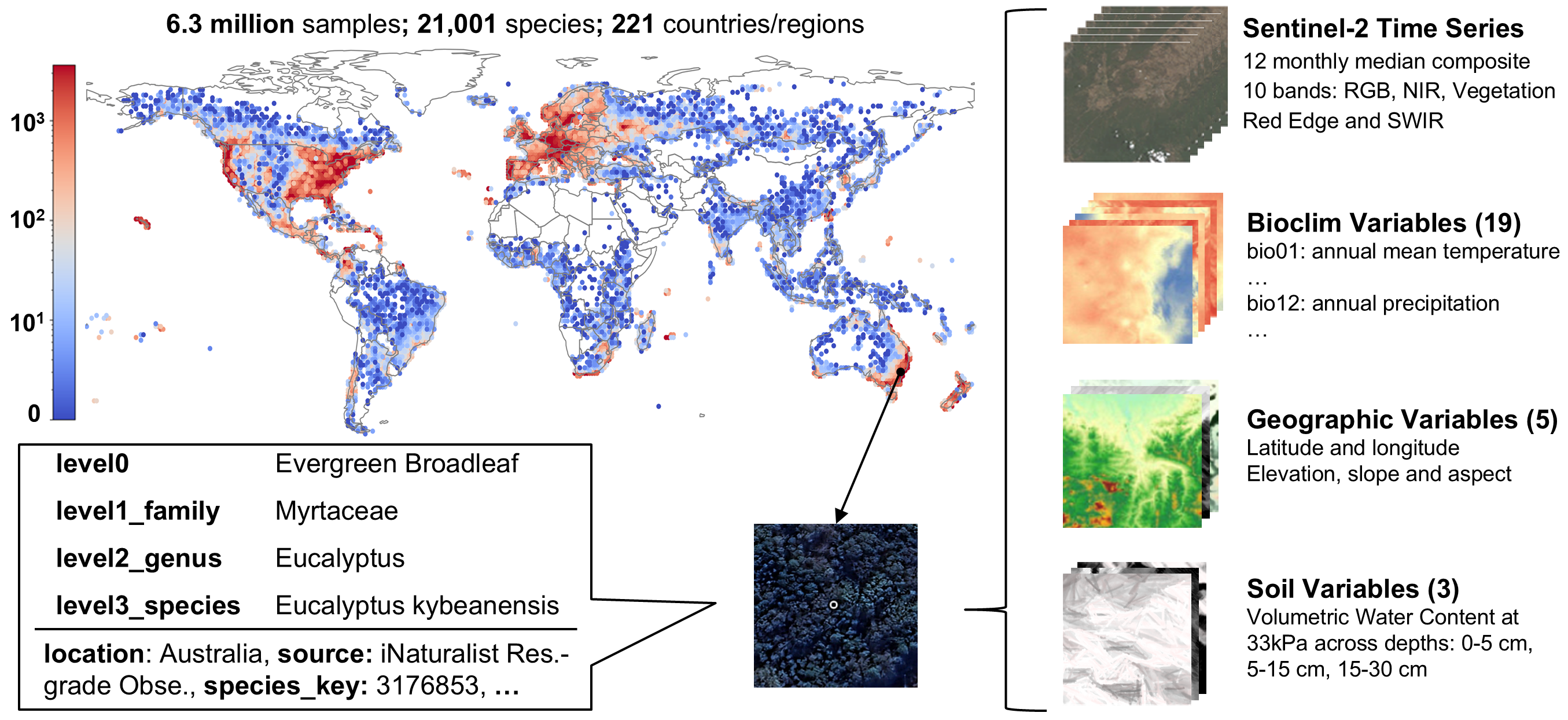}
\caption{Overview of the GlobalGeoTree dataset, which includes 6.3 million samples spanning 21,001 tree species across 221 countries/regions. The map illustrates the geographic coverage, with color intensity representing the number of samples in each 1° × 1° latitude/longitude grid. Each sample is paired with remote sensing data, including Sentinel-2 time series, auxiliary environmental variables, and hierarchical taxonomic labels spanning from functional type to species level.}
\label{fig:dataset_overview}
\end{figure}

Traditional ground-based forest monitoring methods \citep{wellbrock2018leitfaden}, while providing detailed information, are often limited in their spatial and temporal coverage, making it challenging to obtain a comprehensive understanding of global forest composition and dynamics. In contrast, remote sensing has emerged as a key technology for large-scale forest monitoring, offering non-invasive and cost-effective approaches to tree species classification \citep{hermosilla2022mapping}. Despite significant advancements in this field, progress has been constrained by the limited availability of comprehensive, high-quality, and accurately labeled datasets that capture the global diversity of tree species \citep{bountos2025fomo}. Existing datasets typically focus on specific geographic regions or limited taxonomic coverage, hampering the development of models with global applicability \citep{ouaknine2025openforest}.

To bridge these gaps, we present GlobalGeoTree, a large-scale dataset comprising 6.3 million remote sensing samples paired with multi-level taxonomic labels. This dataset integrates time-series satellite imagery from Sentinel-2 with 27 bioclimatic, geographic, and soil variables, offering a rich multimodal representation of tree species within their environmental contexts. The taxonomic hierarchy spans family, genus, and species levels, enabling classification across various scales of biological organization.

In addition to the dataset, we introduce GeoTreeCLIP, a vision-language model specifically designed for tree species classification. Drawing on frameworks like CLIP \citep{radford2021learning}, our approach aligns satellite imagery with taxonomic labels to learn nuanced representations. Unlike traditional classifiers treating labels as discrete categories, vision-language models can inherently process label hierarchical structure, generalizing to unseen species through representations of related genera or families \citep{stevens2024bioclip}. This enables robust zero-shot and few-shot learning of GeoTreeCLIP, which are critical for addressing the vast scale of global biodiversity, ever-evolving species catalogs, and the practical impossibility of exhaustive data collection for all taxa.

GeoTreeCLIP leverages domain-specific pretraining on \textit{GlobalGeoTree-6M}, main part of the dataset tailored for model pretraining, and evaluated on a specialized benchmark, \textit{GlobalGeoTree-10kEval}, which enables a comprehensive assessment of model performance across multiple taxonomic levels. Through the open availability of GlobalGeoTree, its associated models, and evaluation protocols, we seek to establish a community-driven benchmarking standard that will accelerate the development of generalizable models for tree species mapping and deepen our understanding of global forest biodiversity.

\section{Related Work}
\subsection{Open datasets for tree species classification}
Table \ref{tab:tree-datasets} provides an overview of notable open datasets that have contributed to tree species classification, detailing their geographic coverage, size, and taxonomic diversity. For instance, the Seu Nico Forest dataset \citep{gastauer2015tree} from Brazil provides geolocated samples for 228 species but is geographically constrained. Similarly, the Maraca Ecological Station dataset \citep{farias2020dataset} includes 110 species but is also region-specific. In Europe, the EUForest dataset \citep{mauri2017eu} offers broader coverage with data for 242 species. On a global scale, datasets such as Tallo \citep{jucker2022tallo} provide significant taxonomic diversity, covering 5,163 species across 187 families. However, these datasets lack integration with remote sensing or environmental variables, limiting their application in ecological modeling.

\begin{table}[ht]
  \caption{Overview of publicly available datasets for tree species classification.}
  \label{tab:tree-datasets}
  \centering
  \small
  \begin{tabular}{lllll}
    \toprule
    \textbf{Dataset} & \textbf{Geographic scope} & \textbf{Size} & \textbf{Classes} & \textbf{Year} \\
    \midrule
    Seu Nico Forest \citep{gastauer2015tree} & Brazil & 2,868 & 54 families; 139 genera; 228 species & 2015 \\
    EUForest \citep{mauri2017eu} & Europe & 588,983 & 83 genera; 242 species & 2017 \\
    Maraca Eco. Sta. \citep{farias2020dataset} & Brazil & 680 & 40 families; 110 species & 2020 \\
    TreeSatAI \citep{ahlswede2022treesatai} & Germany & 50,381 & 15 genera; 20 species & 2022 \\
    Tallo \citep{jucker2022tallo} & Global & 498,839 & 187 families; 1,453 genera; 5,163 species & 2022 \\
    Indi. Tree Point Clouds \citep{weiser2022individual} & Germany & 1,491 & 22 species & 2022 \\
    NEON Veg. Struc. \citep{kampe2010neon} & USA & N/A & 949 genera; 2,826 species & 2023 \\
    PureForest \citep{gaydon2025pureforest} & France & 135,569 & 18 species & 2024 \\
    Planted \citep{pazos2024planted}  & Global & 2,264,747 & 46 genera; 40 species & 2024 \\
    \midrule
    \textbf{GlobalGeoTree} & \textbf{Global} & \textbf{6,263,345} & \textbf{275 families; 2,734 genera; 21,001 species} & \textbf{2025} \\
    \bottomrule
  \end{tabular}%
\end{table}

Advances in high-resolution imaging and lidar technologies have enabled datasets like PureForest \citep{gaydon2025pureforest} and Individual Tree Point Clouds \citep{weiser2022individual}, which utilize aerial and point cloud data for species classification. While these datasets offer detailed structural information, they remain region-specific and lack the spectral and temporal depth of satellite-based datasets. The TreeSatAI dataset \citep{ahlswede2022treesatai} combines multi-sensor data, including aerial imagery and Sentinel-1/2, for tree species classification in Germany but covers only 20 species. Similarly, the Planted dataset \citep{pazos2024planted} focuses on only 40 planted species globally, limiting its broader applicability. Our work also complements broader biodiversity benchmarks like GeoLifeCLEF \citep{botella2025overview} and GeoPlant \citep{picek2024geoplant}. While these focus on regional, multi-lifeform plant prediction, GlobalGeoTree is the first dataset with a global scope specifically curated for over 21,000 forest tree species.

Collectively, while valuable, these existing datasets highlight the persistent need for a benchmark that synergizes global coverage, deep taxonomic information for forest tree species, and multimodal remote sensing data, a gap GlobalGeoTree aims to fill.

\subsection{Vision-language models for remote sensing applications}
Vision-language models (VLMs) enable the integration of visual and textual information. Among these, Contrastive Language-Image Pretraining (CLIP) \citep{radford2021learning} has demonstrated exceptional zero-shot transfer capabilities by jointly training image and text encoders through a contrastive learning objective, aligning image-text pairs within a shared embedding space. For tree species classification, CLIP's ability to learn from image-text pairings (such as satellite imagery and taxonomic labels) offers a path to capture complex visual and semantic relationships. Its proven zero-shot capabilities are particularly suited for addressing the challenges of identifying species within dynamic and evolving catalogs \citep{stevens2024bioclip}. Meanwhile, its few-shot capabilities tackle the issue of limited labeled data, a common obstacle in biodiversity research, offering an advantage over traditional supervised methods.

In remote sensing, VLMs have been applied to various tasks, including image classification, retrieval, and scene understanding, with domain-specific adaptations yielding significant improvements. For example, RemoteCLIP \citep{liu2024remoteclip}, the first VLM specifically tailored for remote sensing, leverages pretraining on large-scale remote sensing imagery paired with aligned text, achieving state-of-the-art performance in zero-shot classification, linear probing, and few-shot learning. Similarly, SkyCLIP \citep{wang2024skyscript}, SkyCLIP-50 \citep{wang2024skyscript} and CLIP-laion-RS \citep{he2024visual} extend the capabilities of CLIP through continual pretraining on semantically diverse remote sensing image-text pairs. These models demonstrate enhanced generalization and transferability, achieving substantial gains in tasks such as zero-shot scene classification, fine-grained classification, and cross-modal retrieval compared to the original CLIP model.

These advancements underscore the importance of domain-specific pretraining in adapting VLMs for remote sensing applications. Aligning models more closely with the unique characteristics of remote sensing tasks has demonstrated significant potential to advance progress in this field.

\section{The GlobalGeoTree dataset}
\subsection{Geolocated data collection and preprocessing}
The GlobalGeoTree dataset provides unprecedented global and taxonomic coverage for tree species classification using remote sensing data, and the collection involved several key steps:

\subsubsection{Tree species catalog construction}
We constructed a comprehensive tree species catalog by integrating two major global repositories: TreeGOER \citep{kindt2023treegoer} and GlobalTreeSearch \citep{beech2017globaltreesearch}, containing 48,129 and 57,681 tree species respectively. This compilation was further enriched with multiple open-source datasets documented in Table \ref{tab:tree-datasets}. The taxonomic framework was subsequently validated and expanded using the Global Biodiversity Information Facility (GBIF) Species API \citep{gbif_species_api}, ensuring nomenclatural consistency and accuracy. The resulting catalog encompasses 87,845 species, representing the global diversity of tree species.

\subsubsection{Geolocation sampling}
\label{filtering}
For each tree species in our catalog, we queried the GBIF Occurrence API \citep{chamberlain2017r} to retrieve global geolocations with documented occurrences. To ensure data quality and reliability, we applied strict filtering criteria, including: (1) selecting only recent observations recorded between 2015 and 2024; (2) limiting data to human observation records; (3) excluding records with geospatial issues as flagged by GBIF (e.g., country-coordinate mismatches); (4) filtering for occurrences with a "present" status; and (5) removing duplicate entries and observations with low geographic precision. Additionally, we ensured all samples conform to open data licenses (CC0 1.0, CC-BY-4.0, etc.), maintaining the dataset's accessibility and reusability for the broader research community.

\subsubsection{Forest layer filtering}
To ensure that each geolocated observation corresponds to a valid tree, we performed an additional forest cover verification step. \textcolor{blue}{We utilized the EC JRC Global Map of Forest Cover 2020 (Version 2) \citep{bourgoin2025global}}, which has a 10m spatial resolution. Each geolocated point from GBIF was cross-referenced with this map, and only samples located within forest areas were retained. This filtering not only served as an essential quality control measure to enhance data reliability but also defined 2020 as the target year for our study. This allowed us to subsequently acquire the Sentinel-2 time series for 2020, ensuring a direct temporal correspondence between the verified ground observations and the satellite imagery. \textcolor{blue}{While we utilized the binary JRC mask for this study, future operational pipelines could integrate multi-class land cover products such as GLC\_FCS10 \citep{zhang2025glc_fcs10} to constrain candidate species based on mapped forest types.}

\subsection{Dataset Composition and Structure}
\begin{figure}[h!]
\centering
\includegraphics[width=0.85\textwidth]{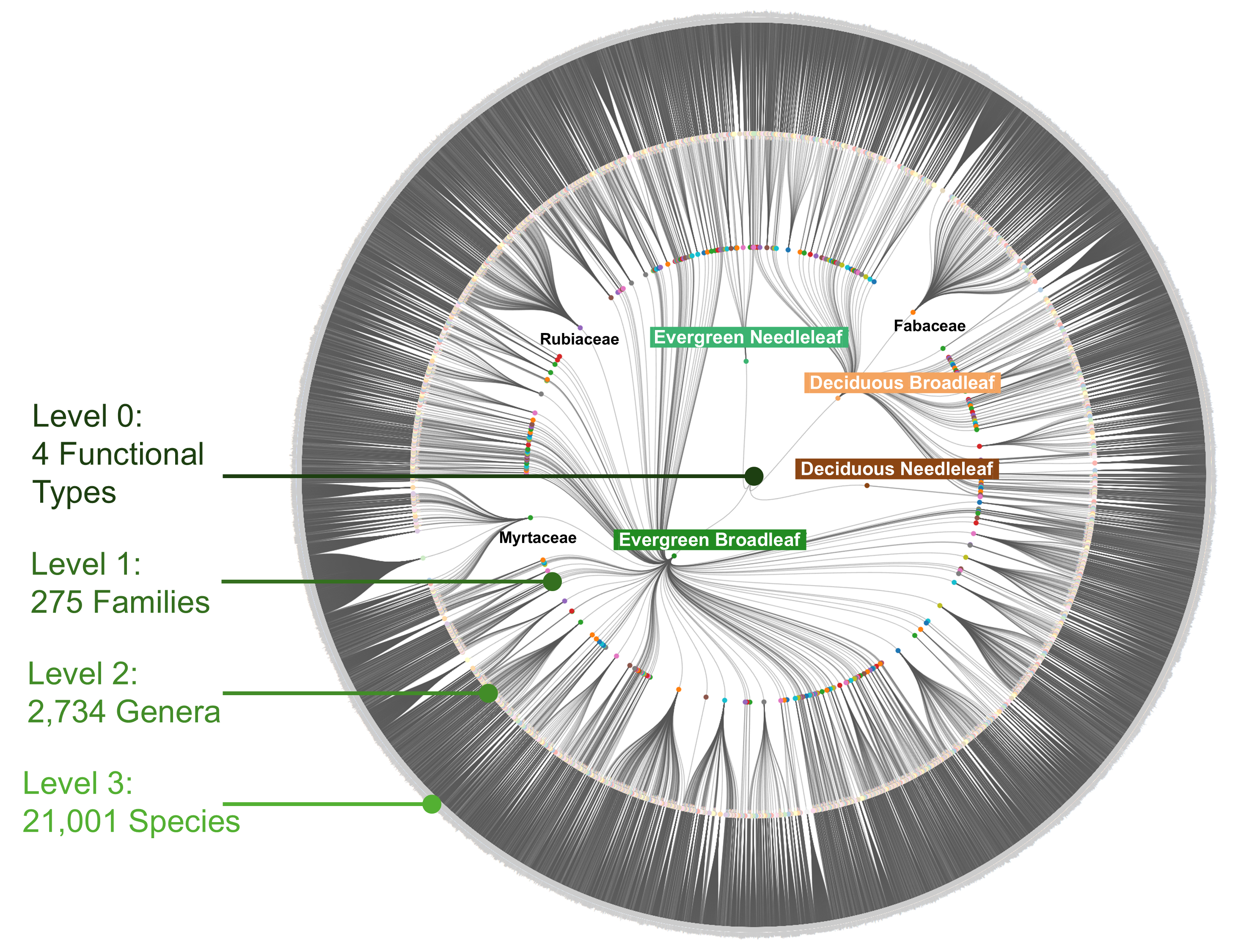}
\caption{The taxonomic hierarchy of the GlobalGeoTree dataset. The visualization shows the nested relationships, branching from the four functional types (e.g., \textit{Deciduous Broadleaf}) at the center, through families (e.g., \textit{Fabaceae}), genera, and out to the 21,001 species at the outermost ring. This multi-level structure is a core feature that enables multi-granular classification tasks.}
\label{fig:taxonomic_hierarchy}
\end{figure}

The filtering process results in the final GlobalGeoTree dataset, which comprises 6,263,345 high-quality samples distributed across 221 countries and regions. \textcolor{blue}{Approximately 70\% of the records are sourced from "Research Grade" iNaturalist observations, which require identification consensus from at least two independent contributors. Other samples are mainly from authoritative datasets available through GBIF, including National Forest Inventory (NFI) data from countries such as Sweden, France, and Colombia. The full list of source datasets and their provenance is traceable via the unique GBIF Derived Dataset DOI of GlobalGeoTree: \url{https://doi.org/10.15468/dd.9qxqyy}.} A core feature of this dataset is its multi-granular taxonomic hierarchy, which is essential for developing models that can generalize across different levels of biological classification. This hierarchical structure is visualized in Fig.~\ref{fig:taxonomic_hierarchy}, illustrating the relationships branching from four broad functional types at the center (level 0), through 275 families (level 1) and 2,734 genera (level 2), out to the 21,001 individual species at the periphery (level 3).

Each of the 6.3 million samples is structured as a single record that includes this full taxonomic lineage, its geolocation, and source metadata. This entire collection is made available as a downloadable CSV file (\texttt{GlobalGeoTree.csv}) for full transparency and accessibility. Table~\ref{tab:sample_records} presents several example records to illustrate the data structure. A comprehensive statistical breakdown and accessibility details of the dataset is provided in Appendix~\ref{app:ggt_stats}.

\begin{table}[h!]
\centering
\caption{Example records from the \texttt{GlobalGeoTree.csv} file, demonstrating the structure of each geolocated data point.}
\label{tab:sample_records}
\resizebox{\textwidth}{!}{%
\begin{tabular}{llllllll}
\toprule
    \textbf{country\_code} & \textbf{level0} & \textbf{level1\_family} & \textbf{level2\_genus} & \textbf{level3\_species} & \textbf{source} & \textbf{longitude} & \textbf{latitude} \\
    \midrule
    ES & Deciduous Broadleaf & Fagaceae & \textit{Quercus} & \textit{Quercus rotundifolia} & iNaturalist Res.-grade Obs. & -3.74024 & 40.49546 \\
    US & Evergreen Needleleaf & Cupressaceae & \textit{Thuja} & \textit{Thuja plicata} & iNaturalist Res.-grade Obs. & -122.168 & 48.33392 \\
    IE & Deciduous Broadleaf & Fagaceae & \textit{Castanea} & \textit{Castanea sativa} & iNaturalist Res.-grade Obs. & -9.524 & 52.05556 \\
    PT & Deciduous Broadleaf & Fagaceae & \textit{Quercus} & \textit{Quercus rotundifolia} & Forestry Inventory 2015 & -7.85145 & 39.42701 \\
    AU & Deciduous Broadleaf & Fabaceae & \textit{Acacia} & \textit{Acacia platycarpa} & ALA species sightings & 132.9895 & -14.818 \\
\bottomrule
\end{tabular}%
}
\end{table}

\subsection{Paired remote sensing data}
Each sample in the GlobalGeoTree dataset is paired with a rich set of multimodal data to enable robust modeling. Table~\ref{tab:globalgeotree-features} provides a comprehensive overview of these features, including paired Earth Observation (EO) data and auxiliary environmental variables derived from remote sensing sources.

\begin{table}[ht]
  \caption{Overview of features in each sample in the GlobalGeoTree dataset.}
  \label{tab:globalgeotree-features}
  \centering
  \small
  \begin{tabular}{p{5cm}lp{8cm}}
    \toprule
    \textbf{Feature Name} & \textbf{Type} & \textbf{Description} \\
    \midrule
    \multicolumn{3}{c}{\textbf{Remote Sensing Data}} \\
    \midrule
    Sentinel-2 Time Series & float & 12 monthly median composites; Includes RGB, NIR, Vegetation Red Edge, and SWIR bands; dimensions: (12, 10, 5, 5). \\
    Geographic Variables & float & Latitude and longitude, as well as elevation, slope, and aspect derived from USGS (SRTM) (30m resolution). \\
    Soil Variables & float & 3 Volumetric Water Content data at 33kPa across depths: 0-5 cm, 5-15 cm, 15-30 cm (250m resolution). \\
    Bioclim Variables & float & 19 climatic variables from WorldClim (1km resolution). \\
    \midrule
    \multicolumn{3}{c}{\textbf{Text Labels}} \\
    \midrule
    level0 & string & Functional type of the species (e.g., Evergreen Broadleaf). \\
    level1\_family & string & Taxonomic family of the species (e.g., Myrtaceae). \\
    level2\_genus & string & Taxonomic genus of the species (e.g., \textit{Eucalyptus}). \\
    level3\_species & string & Scientific name of the species (e.g., \textit{Eucalyptus kybeanensis}). \\
    \midrule
    \multicolumn{3}{c}{\textbf{Meta Data}} \\
    \midrule
    location & string & Geographic location of the sample (e.g., Australia). \\
    country\_code & string & ISO country code of the sample location. \\
    source & string & Source of the sample record (e.g., iNaturalist Research-grade Observations). \\
    species\_key & float & Unique identifier for the species in the GBIF database. \\
    record\_year & int & The year when the record was collected. \\
    \bottomrule
  \end{tabular}%
\end{table}

\subsubsection{EO data}
The Sentinel-2 data for each sample consists of a time series of 12 monthly median composites from January to December 2020. The full-year temporal coverage enables models to capture the distinct seasonal patterns of different tree species, such as leaf-out, full canopy, and senescence \citep{mu2025national}. These phenological features are particularly discriminative for deciduous species.

For each month, all L2A Sentinel-2 images with less than 30\% cloud cover were collected, and the median composites were generated from these images. This process ensures a robust, largely cloud-free representation for each month, effectively mitigating issues caused by transient atmospheric conditions in single-date imagery.

Each composite includes a 5×5 pixel patch centered on the geolocation of the tree species. This patch size was chosen to account for typical crown sizes (10–30 m) \citep{jucker2022tallo} and to align with other public datasets, such as PureForest \citep{gaydon2025pureforest} and TreeSatAI \citep{ahlswede2022treesatai}. The chosen 5×5 patch size represents a trade-off: it is sufficiently large enough to capture the crown of a mature tree and its immediate context, while remaining compact enough to minimize noise from surrounding unrelated species.

\subsubsection{Auxiliary data}
The auxiliary data enriches the EO data for each sample by providing additional contextual environmental information. Geographic variables, such as elevation, slope, and aspect, are derived from the USGS SRTM dataset \citep{jarvis2008hole}, while soil data, including volumetric water content at 33 kPa across three depths (0–5 cm, 5–15 cm, 15–30 cm), are obtained from SoilGrid \citep{poggio2021soilgrids}. Additionally, 19 bioclimatic variables are sourced from WorldClim \citep{fick2017worldclim}. Due to the relatively coarse spatial resolution of these datasets (ranging from 30m to 1km), only the values corresponding to the exact coordinates of each occurrence are extracted to ensure precision and relevance \citep{gillespie2024deep}. A detailed description of all 27 auxiliary variables and their sources is provided in Appendix \ref{app:aux_env}.

\subsection{Dataset partitioning}
\label{3.3}
For effective model development and evaluation, the GlobalGeoTree dataset was partitioned into \textit{GlobalGeoTree-6M} and curated
evaluation subsets, primarily \textit{GlobalGeoTree-10kEval}.

\begin{figure}[h]
  \centering
  \includegraphics[width=0.7\textwidth]{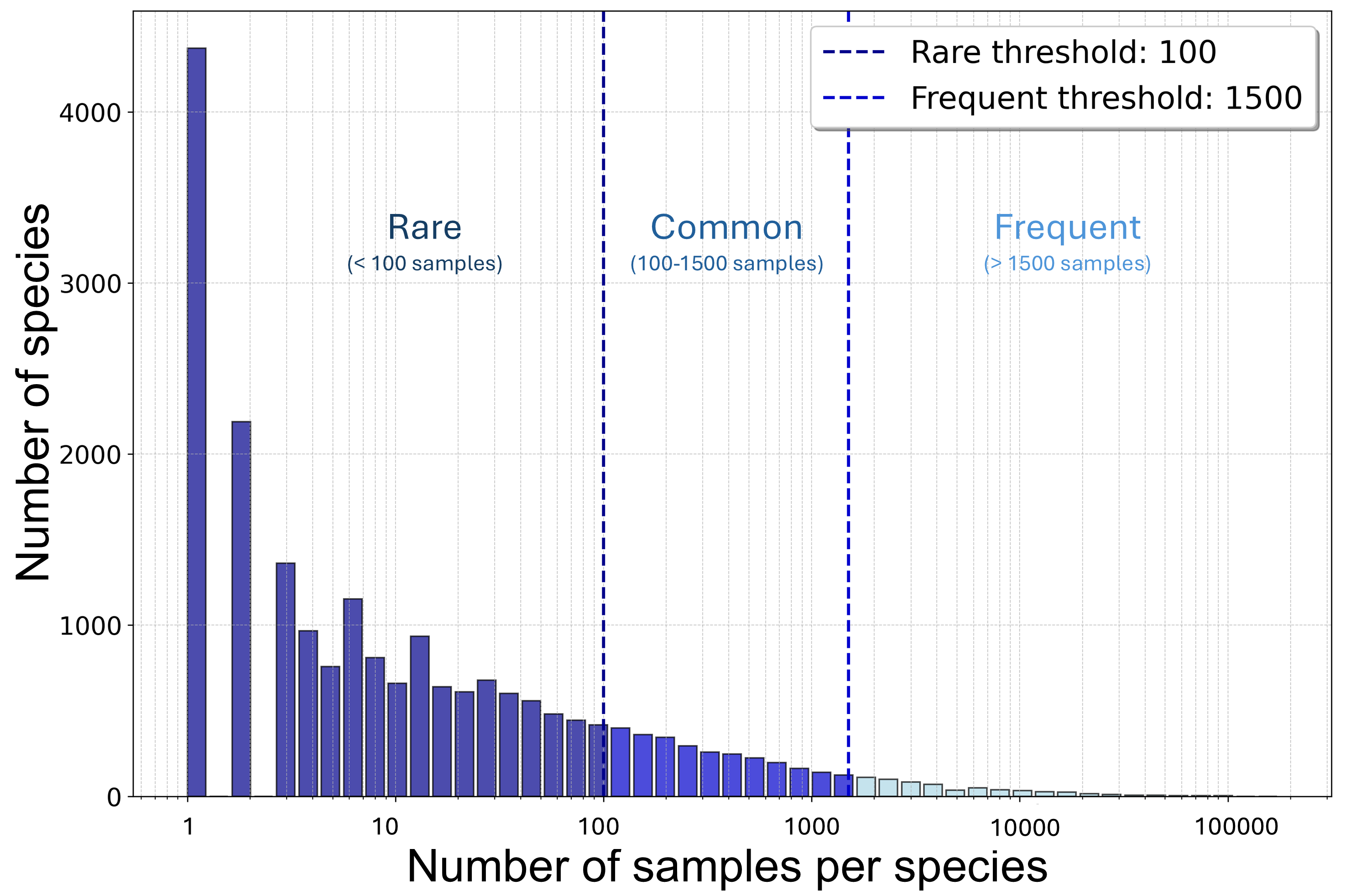}
  \caption{Species in GlobalGeoTree are categorized into Frequent, Common and Rare groups based on the number of samples per species.}
  \label{fig:species_distribution}
\end{figure}

\textit{GlobalGeoTree-6M} comprises the vast majority of the samples and is specifically designed for model pretraining. This large size allows models to learn robust and generalizable representations of tree species and their associated environmental contexts.

\textcolor{blue}{\textit{GlobalGeoTree-10kEval} is a carefully curated dataset designed to benchmark model performance across taxonomic levels and species frequency categories. To address the characteristic long-tail distribution (detailed in Appendix \ref{longtail}), we categorized species into three groups based on sample frequency: Frequent ($>1500$ samples), Common ($100\text{--}1500$ samples), and Rare ($<100$ samples), as shown in Fig. \ref{fig:species_distribution}.}

\begin{figure}[h]
\centering
\includegraphics[width=0.75\textwidth]{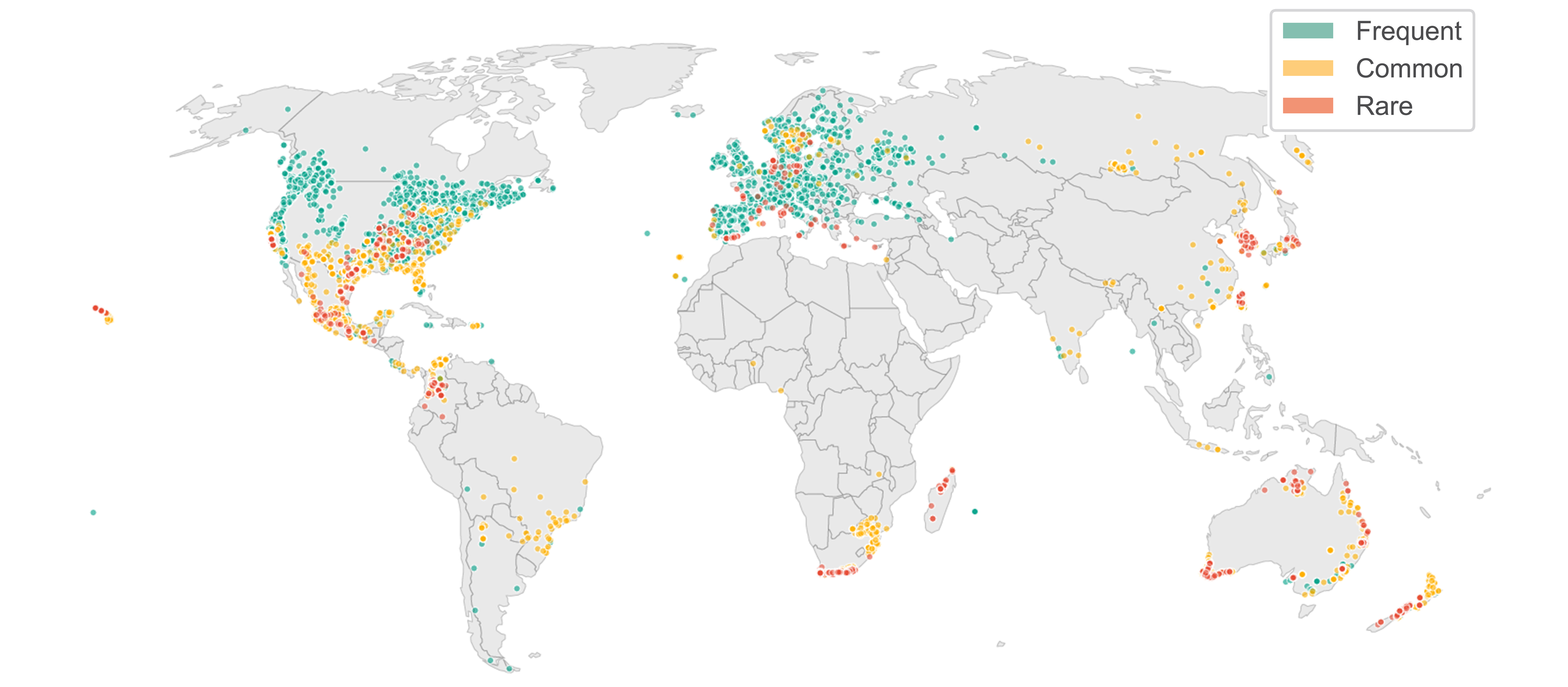}
\caption{Geographic distribution of \textit{GlobalGeoTree-10kEval}. This benchmark includes species selected from Frequent, Common, and Rare categories, as described in the text.}
\label{fig:10kEval_distribution}
\end{figure}

\textcolor{blue}{The primary \textit{GlobalGeoTree-10kEval} dataset includes 30 species from each of these three categories, resulting in a total of 90 species. The sample proportions within this evaluation set are 12\% for Rare species, 33\% for Common species, and 55\% for Frequent species, culminating in around 10,000 samples.} Fig. \ref{fig:10kEval_distribution} shows the geographical distribution of \textit{GlobalGeoTree-10kEval}, which spans diverse regions across the globe. This global distribution ensures that the dataset captures a wide range of ecological and environmental contexts, making it representative of real-world scenarios. By focusing on a diverse set of species with varying levels of representation, \textit{GlobalGeoTree-10kEval} serves as a robust evaluation benchmark for assessing the ability of models to tackle challenges posed by the long-tail distribution of species and shifts in geographical domains.

To further evaluate model robustness and scalability across broader taxonomic scopes, \textcolor{blue}{we constructed two additional evaluation subsets: \textit{GlobalGeoTree-10kEval-300} and \textit{GlobalGeoTree-10kEval-900}, containing 100 and 300 species per category, respectively. Crucially, all samples within these evaluation sets, were subjected to the rigorous filtering criteria detailed in Sect. \ref{filtering} and are strictly excluded from the \textit{GlobalGeoTree-6M} pretraining set to ensure fair evaluation.} Details of all evaluation subsets (Appendix \ref{eval_details}) and the corresponding evaluation results (Appendix \ref{performance_eval}) are also provided. Given the complexity of global tree species classification, our primary analysis focuses on the 90-species \textit{GlobalGeoTree-10kEval}, which serves as a practical starting point for systematic benchmarking.

\color{blue}
\section{Data Quality and Validation}
\label{sec:validation}

\begin{figure}[ht]
\centering
\includegraphics[width=\textwidth]{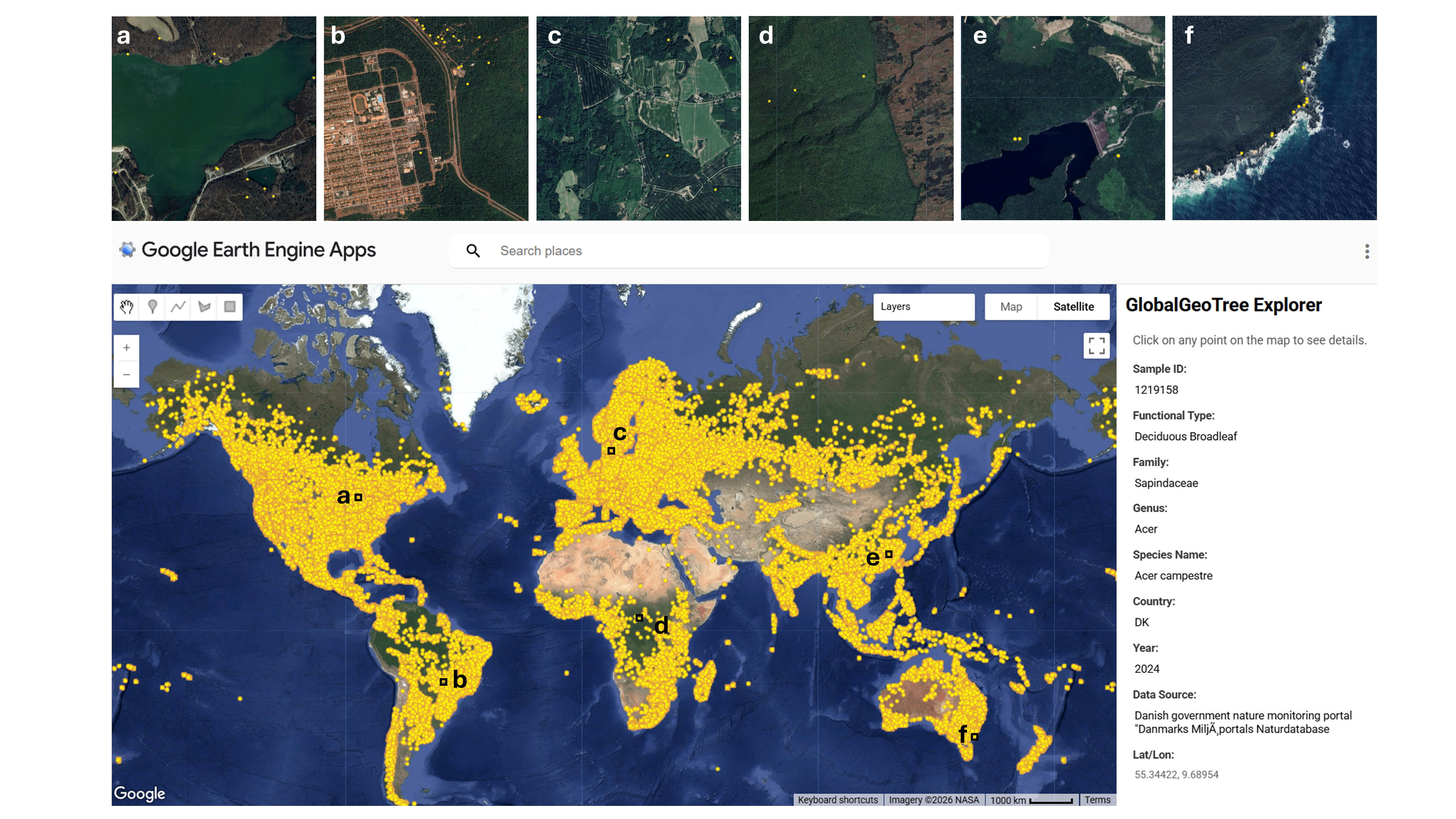}
\caption{\color{blue}The GlobalGeoTree Explorer App. The interface allows users to visualize the global distribution of tree samples (bottom map) and inspect individual points against high-resolution satellite imagery (top panels a-f). Detailed attributes for selected samples are displayed in the sidebar, facilitating transparent community validation.}
\label{fig:gee_app}
\end{figure}

To ensure the reliability of GlobalGeoTree, we implemented a multi-tiered validation framework combining fine-scale visual inspection, global-scale cross-verification with independent land cover products, and a public platform for community auditing.

\subsection{Visual Inspection and Community Validation}
We first quantified land-cover accuracy through a visual validation study. A random subset of 300 locations was sampled from \textit{GlobalGeoTree-10kEval}. For each sample, we retrieved Very High Resolution (VHR) satellite imagery to verify whether the geolocated point fell within forest or tree cover. The inspection confirmed that 98.3\% of the samples were correctly located in forested areas, while a small minority were ambiguous (e.g., forest edges), validating the efficacy of our automated filtering pipeline.

To further enhance transparency and enable large-scale qualitative validation by the research community, we developed the GlobalGeoTree Explorer, a web-based application hosted on Google Earth Engine (Fig.~\ref{fig:gee_app}). This tool allows users to visualize the spatial distribution of all 6.3 million samples, overlay them on high-resolution satellite basemaps, and inspect detailed taxonomic attributes for any individual point. The application is publicly accessible at \url{https://ee-yangm.projects.earthengine.app/view/globalgeotree-explorer}.

\subsection{Cross-Validation with Global Land Cover Products}
To validate taxonomic consistency at the functional type level, we performed a cross-comparison against the Copernicus Global Land Cover Layers (CGLS-LC100, Collection 3) \citep{buchhorn2020copernicus}. Since no global dataset currently provides high-resolution distribution maps at the species or genus level, the CGLS-LC100 product represents the most detailed global baseline available, offering discrete land cover layers at 100m resolution that distinguish between broadleaf and needleleaf forest types.

\begin{figure}[ht]
\centering
\includegraphics[width=\textwidth]{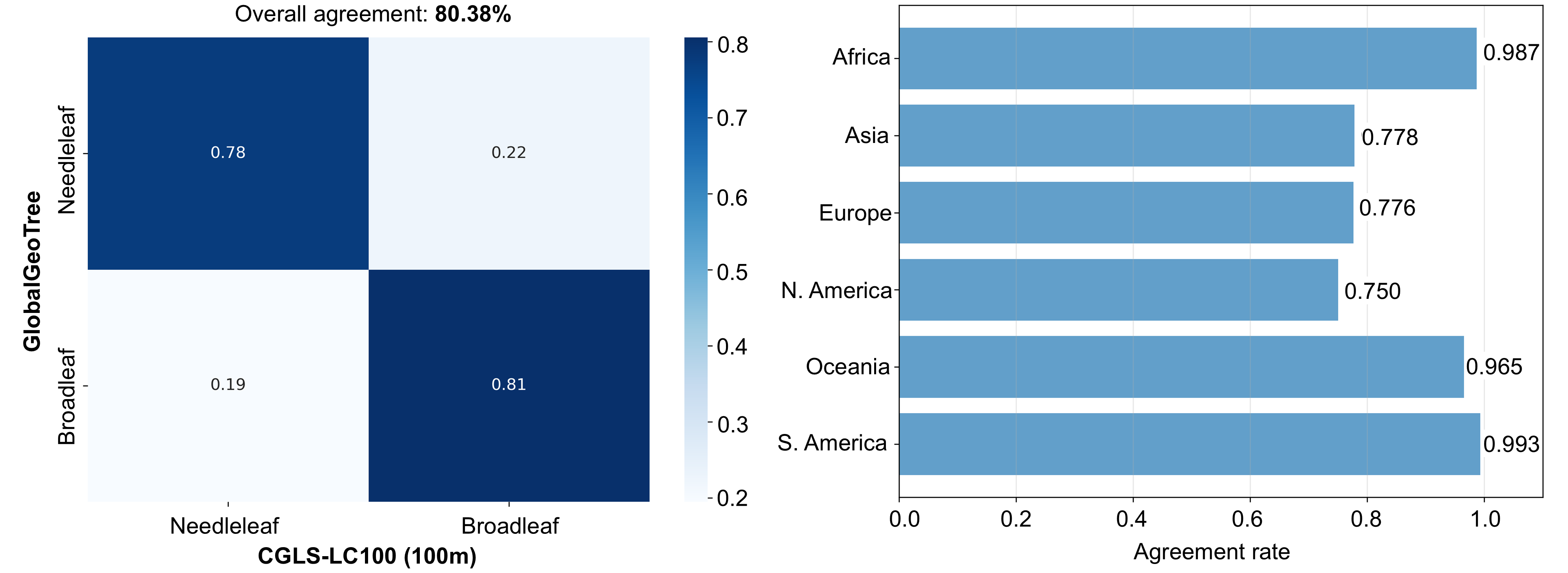}
\caption{\color{blue}Cross-validation results against the CGLS-LC100 product. \textbf{Left:} Normalized confusion matrix at the Leaf Level (Needleleaf vs. Broadleaf), showing an overall agreement of 80.38\%. \textbf{Right:} Agreement rates broken down by continent, highlighting strong consistency in tropical regions.}
\label{fig:cross_val}
\end{figure}

We aligned the \texttt{level0} labels of GlobalGeoTree with CGLS classes by aggregating them into two primary leaf types: Needleleaf and Broadleaf. To ensure global representativeness and mitigate regional data density biases, we employed a stratified sampling strategy, selecting 10,000 samples per continent. 

Figure~\ref{fig:cross_val} presents the results of this cross-validation. The confusion matrix (left) reveals an overall agreement of 80.38\% at the leaf level. We observed high consistency for Broadleaf samples (81\%), while some Needleleaf samples (22\%) showed discrepancies with the coarser 100m product, likely due to mixed pixels in transition zones. The regional breakdown (right) demonstrates exceptionally high agreement in South America (99.3\%) and Africa (98.7\%), where broadleaf forests dominate. Lower agreement rates in North America and Europe are expected given the higher prevalence of mixed forests and the resolution gap between our point-based data and the 100m validation raster.

\subsection{Taxonomic Reliability}
Beyond spatial validation, taxonomic reliability is ensured through our strict data provenance pipeline. We exclusively utilize observations which require community consensus for species identification, and 100\% of these samples have passed the GBIF taxonomic backbone check \citep{GBIF_Backbone_2023}.

\color{black}
\section{Benchmarks}
\subsection{GeoTreeCLIP model}
\label{4.1}
To establish a strong baseline on the GlobalGeoTree dataset, we developed GeoTreeCLIP, a vision-language model (VLM) specifically tailored for tree species classification. Our choice of the VLM paradigm was motivated by the inherent challenges of this task: preliminary experiments showed that traditional supervised models struggle with the vast, long-tailed, and hierarchical label space of GlobalGeoTree, especially at the fine-grained genus and species levels (see Sect.~\ref{app:supervised_comparison}). 

In contrast, a VLM framework treats taxonomic labels not as discrete indices, but as structured text. This allows the model to leverage the rich semantic relationships between species, genera, and families. By jointly training on image-text pairs, GeoTreeCLIP learns powerful, transferable representations that align visual features with their corresponding taxonomic context. This approach is particularly effective for zero-shot and few-shot scenarios, which are critical for addressing the immense scale of global biodiversity and the practical impossibility of collecting exhaustive data for all taxa \citep{stevens2024bioclip}.

\begin{figure}[h]
\centering
\includegraphics[width=0.85\textwidth]{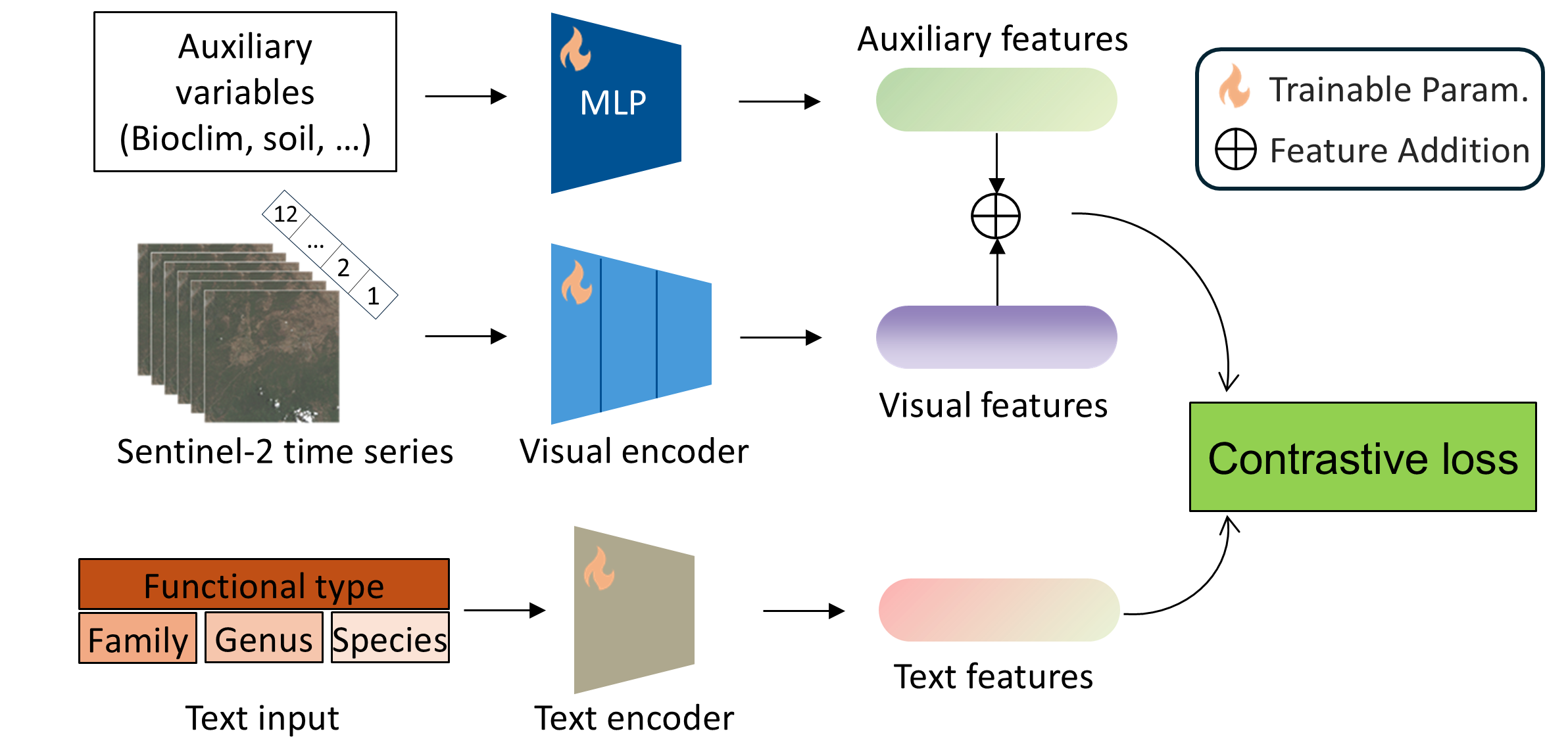}
\caption{Architecture of the GeoTreeCLIP baseline model. It processes Sentinel-2 time series and auxiliary data through visual encoder and MLP, and hierarchical taxonomic labels through a text encoder. The resulting multimodal visual and text features are then aligned using a contrastive loss.}
\label{fig:model}
\end{figure}

As shown in Fig. \ref{fig:model}, the GeoTreeCLIP model architecture consists of the following components:
\begin{itemize}
    \item \textbf{Visual Encoder}:  A ViT-B/16 backbone \citep{dosovitskiy2020image} adapted to process the multi-spectral, time-series Sentinel-2 data. To handle the 10-channel, 5x5 pixel input patches, we modify the initial patch embedding layer of the ViT. Each 5x5 patch is first upsampled to 16x16 pixels using bicubic interpolation. The modified embedding layer then processes these 10 channels and projects the entire 16x16 patch into a single visual token. This approach allows the model to capture the holistic spectral information of the patch at each timestep. These 12 monthly tokens are then fed into a temporal attention module \citep{vaswani2017attention} to learn and fuse key phenological patterns across the year. This mechanism allows the model to dynamically weigh the importance of different months, implicitly learning which parts of the phenological cycle are most informative for identifying a particular species.
    \item \textbf{Auxiliary Feature Integration}: A multi-layer perceptron (MLP) \citep{rosenblatt1958perceptron, gillespie2024deep} designed to process bioclimatic, soil, and geographic data. The MLP consists of several feature-specific linear layers (hidden dimension: 256) followed by LayerNorm and ReLU activation. The encoded features are then fused and projected into a 768-dimensional embedding, which is added to the visual token embedding from the Visual Encoder before the final contrastive projection. This allows the model to integrate environmental context with visual information. \textcolor{blue}{We adopt a dual-branch architecture comprising a Visual Encoder and a separate MLP for auxiliary features, rather than stacking all inputs into a single data cube. This design addresses the fundamental heterogeneity of the data modalities. First, Sentinel-2 data consists of a dynamic 12-month time series capturing phenology, whereas auxiliary variables such as bioclimatic and topographic metrics are static. Integrating these static variables into the visual branch would require replicating them across the temporal dimension, introducing significant data redundancy. Second, the spatial resolution of auxiliary variables is typically much coarser ($\sim$1km or 250m) than the 10m Sentinel-2 imagery. Treating them as image bands would necessitate upsampling, creating spatially invariant feature maps that are computationally inefficient for the Vision Transformer. By processing these modalities separately, the MLP branch efficiently encodes environmental context as a conditional prior, which is subsequently fused with spatiotemporal visual features in the shared embedding space.}
    \item \textbf{Text Encoder}: A 77-token causal autoregressive transformer \citep{vaswani2017attention, radford2019language} that processes the hierarchical taxonomic labels. For each sample, the text input is structured to convey the full taxonomic lineage, for example, "Evergreen Broadleaf, Family Myrtaceae, Genus Eucalyptus, Species Eucalyptus kybeanensis". By encoding this structured text, the model learns embeddings that capture the relationships between different taxonomic levels. This allows GeoTreeCLIP to understand, for instance, that two different species within the same genus (\textit{Eucalyptus}) are semantically closer than species from different families. This learned semantic structure is key to its strong generalization capabilities, especially for rare or unseen species.
\end{itemize}

\subsubsection{Pretraining details}
Both the Visual Encoder and Text Encoder are initialized using OpenAI’s publicly available CLIP checkpoint \citep{radford2021learning} and further pre-trained on the \textit{GlobalGeoTree-6M} using Distributed Data Parallel (DDP) across 5 NVIDIA 3090 (24GB) GPUs. We employed a batch size of 384 per GPU, with gradient accumulation over 2 steps, resulting in an effective batch size of 768 per GPU (3840 globally). The AdamW optimizer \citep{loshchilov2017decoupled} was used with a base learning rate of $1 \times 10^{-5}$ for the visual and auxiliary encoders, and a reduced learning rate of $1 \times 10^{-6}$ for the pretrained text encoder. A weight decay of $1 \times 10^{-4}$ was applied. A linear learning rate warmup was implemented for the first 5 epochs. Following the warmup, a Cosine Annealing with Warm Restarts \citep{loshchilov2022sgdr} scheduler was used, with $T_0=10$ epochs and $T_{mult}=2$, and a minimum learning rate of $1 \times 10^{-7}$. Gradients were clipped to a maximum L2 norm of 1.0. The loss function was the standard CLIP contrastive loss \citep{oord2018representation}. Mixed-precision training \citep{micikevicius2017mixed} was enabled. A full 25-epoch training run required approximately 2 days, with peak GPU memory consumption observed at roughly 14 GB per GPU.

\subsection{Experimental setup}
\label{4.2}
We evaluated GeoTreeCLIP against two advanced pretrained vision-language models: CLIP \citep{radford2021learning} and RemoteCLIP \citep{liu2024remoteclip}, a specialized VLM for remote sensing applications. To ensure a fair comparison with models not explicitly designed for time-series data, we employed an ensemble-like approach for CLIP and RemoteCLIP. Specifically, features or probabilities were computed independently for each of the 12 monthly images, and the final results were obtained by averaging these values. Additionally, the results of other baseline models, including SkyCLIP-50 \citep{wang2024skyscript} and CLIP-laion-RS \citep{he2024visual}, are provided in Appendix \ref{app:additional_zeroshot_baselines_eval10k}.

All models were evaluated on the \textit{GlobalGeoTree-10kEval} benchmark using zero-shot and few-shot learning settings. Performance was measured using top-1 and top-5 prediction accuracy, with separate evaluations for each taxonomic level (family, genus, and species). To ensure robustness, we repeated each experiment 5 times using different random seeds and reported the mean accuracy and variance.

\subsubsection{Zero-shot evaluation}
For zero-shot evaluation, we assess each model's ability to classify samples from the evaluation sets (\textit{GlobalGeoTree-10kEval}, \textit{-10kEval-300}, and \textit{-10kEval-900}) without any fine-tuning. Our protocol is framed as an "in-domain" zero-shot classification task, following the setup in recent remote sensing benchmarks \citep{wang2024skyscript}. This approach tests a model's transfer capability to new, unseen samples of species, rather than to entirely unseen categories. Although the species categories in the evaluation sets are present in the training data, this setup strikes a balance between preserving the model’s ability to classify all tree species after pre-training on \textit{GlobalGeoTree-6M} and ensuring a valid zero-shot scenario.

\subsubsection{Few-shot evaluation}
For few-shot evaluation, we explored scenarios such as one-shot learning, where the model is provided with only one labeled example per species. To implement this, we adopted a fine-tuning-based approach \citep{parnami2022learning} using the pre-trained model. Specifically, we randomly sampled $k$ labeled examples per class (e.g., $k=1$, $k=3$) to form the support set and fine-tuned the visual encoder of the pre-trained model on this set. During fine-tuning, most of the visual encoder's parameters were frozen, with only the last four transformer layers and the classification-related parameters remaining trainable. The text encoder was entirely frozen, leveraging the pre-trained textual embeddings for class labels. The fine-tuning was conducted for 10 epochs to balance adaptation and prevent overfitting due to the small support set. Afterward, the model was evaluated on the query set, which consisted of the remaining examples in the dataset. 

For each query image, predictions were made by computing similarity scores between its visual embedding (extracted by the fine-tuned visual encoder) and the textual embeddings of the class labels. The class with the highest similarity score was assigned as the predicted label. Classification accuracy on the query set was then used to evaluate the model's performance. This fine-tuning approach enables the model to adapt to the few-shot setting while retaining the benefits of the pre-trained representations.

This evaluation framework highlights the model's ability to generalize effectively from limited labeled data, which is a crucial capability for real-world applications in biodiversity monitoring where obtaining large amounts of labeled data for every species is often infeasible.

\subsection{Experimental results}
\label{4.3}
\subsubsection{Zero-shot evaluation}

\begin{table}[h]
\centering
\caption{Zero-shot evaluation on \textit{GlobalGeoTree-10kEval}. Results are presented as mean accuracy (\%) $\pm$ standard deviation (\%) over 5 runs.}
\label{tab:zero-shot}
\begin{tabular}{lcccccc}
\toprule
\multirow{2}{*}{\textbf{Taxon.}} & \multicolumn{2}{c}{\textbf{CLIP}} & \multicolumn{2}{c}{\textbf{RemoteCLIP}} & \multicolumn{2}{c}{\textbf{GeoTreeCLIP}} \\
\cmidrule(lr){2-3} \cmidrule(lr){4-5} \cmidrule(lr){6-7}
& Top-1 & Top-5 & Top-1 & Top-5 & Top-1 & Top-5 \\
\midrule
Family & 10.80 $\pm$ 0.03 & 25.32 $\pm$ 0.05 & 1.11 $\pm$ 0.01 & 10.55 $\pm$ 0.04 & \textbf{20.99 $\pm$ 0.28} & \textbf{56.88 $\pm$ 0.42} \\
Genus  &  1.09 $\pm$ 0.01 &  9.34 $\pm$ 0.01 & 1.11 $\pm$ 0.01 &  6.25 $\pm$ 0.02 & \textbf{18.39 $\pm$ 0.26} & \textbf{50.98 $\pm$ 0.41} \\
Species&  1.09 $\pm$ 0.01 &  7.02 $\pm$ 0.02 & 1.11 $\pm$ 0.01 &  6.25 $\pm$ 0.02 & \textbf{16.71 $\pm$ 0.25} & \textbf{47.52 $\pm$ 0.37} \\
\bottomrule
\end{tabular}%
\end{table}

The results of zero-shot evaluation are presented in Table \ref{tab:zero-shot}, clearly demonstrating the substantial improvements achieved by GeoTreeCLIP across all taxonomic levels. At the family level, GeoTreeCLIP achieves a top-1 accuracy of 20.99\% and a top-5 accuracy of 56.88\%. The performance gap is even more pronounced at the genus level. GeoTreeCLIP achieves a top-1 accuracy of 18.39\% and a top-5 accuracy of 50.98\%, outperforming CLIP (1.09\% top-1, 9.34\% top-5) and RemoteCLIP (1.11\% top-1, 6.25\% top-5) by a large margin. At the most challenging species level, GeoTreeCLIP still shows significant superiority, achieving a top-1 accuracy of 16.71\% and a top-5 accuracy of 47.52\%.

The experimental results reveal two key patterns. First, accuracy consistently declines as the taxonomic level becomes finer, reflecting the growing challenge of distinguishing closely related classes. This trend is observed across all models but is particularly pronounced for CLIP and RemoteCLIP, which perform poorly at the genus and species levels. In contrast, GeoTreeCLIP demonstrates stronger performance at these fine-grained levels, likely due to its ability to learn and leverage the hierarchical relationships in taxonomic labels, as supported by feature embedding visualizations (see Sect. \ref{app:tsne_visualization} and Figure \ref{fig:tsne_all_models_comparison}).  \textcolor{blue}{Second, the significant performance gap between GeoTreeCLIP and the baseline models underscores synergistic effect of domain-specific pretraining and its tailored architecture. Unlike general-purpose models such as CLIP and RemoteCLIP, our approach effectively leverages spatiotemporal and multispectral information to enhance classification capabilities.} Additional zero-shot benchmark results, including evaluations of SkyCLIP-50 \citep{wang2024skyscript} and CLIP-laion-RS \citep{he2024visual}, are provided in Appendix \ref{app:additional_zeroshot_baselines_eval10k}.

\color{blue}
\subsubsection{Performance Analysis by Species Rarity}
To assess model robustness against the long-tail distribution inherent in global biodiversity data, we analyzed zero-shot performance at the species level across three frequency groups: Rare, Common, and Frequent. Table \ref{tab:rarity_performance} details the mean accuracy and standard deviation over 5 runs on the \textit{GlobalGeoTree-10kEval} benchmark.

\begin{table}[h]
\centering
\caption{\color{blue}Zero-shot species-level performance breakdown by rarity group on \textit{GlobalGeoTree-10kEval}. Results are mean accuracy (\%) $\pm$ standard deviation (\%) over 5 runs.}
\label{tab:rarity_performance}
\begin{tabular}{lcccccc}
\toprule
\multirow{2}{*}{\textbf{Rarity Group}} & \multicolumn{2}{c}{\textbf{CLIP}} & \multicolumn{2}{c}{\textbf{RemoteCLIP}} & \multicolumn{2}{c}{\textbf{GeoTreeCLIP}} \\
\cmidrule(lr){2-3} \cmidrule(lr){4-5} \cmidrule(lr){6-7}
 & Top-1 & Top-5 & Top-1 & Top-5 & Top-1 & Top-5 \\
\midrule
Rare     & 0.00 $\pm$ 0.00 & 3.61 $\pm$ 0.05 & 0.00 $\pm$ 0.00 & 2.43 $\pm$ 0.13 & \textbf{15.25 $\pm$ 0.04} & \textbf{46.65 $\pm$ 0.47} \\
Common   & 0.64 $\pm$ 0.04 & 4.52 $\pm$ 0.03 & 0.00 $\pm$ 0.00 & 3.45 $\pm$ 0.00 & \textbf{15.93 $\pm$ 0.08} & \textbf{47.51 $\pm$ 0.08} \\
Frequent & 3.00 $\pm$ 0.03 & 9.27 $\pm$ 0.02 & 3.35 $\pm$ 0.01 & 12.90 $\pm$ 0.04 & \textbf{17.52 $\pm$ 0.36} & \textbf{51.88 $\pm$ 0.12} \\
\bottomrule
\end{tabular}%
\end{table}

The experimental results indicate substantial differences in model generalization across frequency groups. Both baseline models show significant performance degradation on data-scarce classes. Specifically, CLIP yields 0.00\% top-1 accuracy for the Rare group, while RemoteCLIP records 0.00\% top-1 accuracy across both Rare and Common categories. Conversely, GeoTreeCLIP exhibits consistent performance stability across the frequency spectrum. The model achieves a top-1 accuracy of 15.25\% on Rare species, showing a relatively small decrease compared to the 17.52\% accuracy observed for Frequent species. This limited performance disparity suggests that GeoTreeCLIP facilitates knowledge transfer through hierarchical taxonomic relationships, enabling the identification of rare species by leveraging features learned from more frequent, related taxa.

\subsubsection{Performance Analysis by Geographic Region}
To evaluate model generalizability across different geographical regions, we disaggregated the zero-shot results on \textit{GlobalGeoTree-10kEval} by continent. Table \ref{tab:regional_performance} presents the species-level accuracy for each region.

\begin{table}[h]
\centering
\caption{\color{blue}Zero-shot species-level performance by continent on \textit{GlobalGeoTree-10kEval}. Results are mean accuracy (\%) $\pm$ standard deviation (\%) over 5 runs. (\textit{n} indicates the sample count per region in the evaluation set).}
\label{tab:regional_performance}
\resizebox{\textwidth}{!}{%
\begin{tabular}{lcccccc}
\toprule
\multirow{2}{*}{\textbf{Region}} & \multicolumn{2}{c}{\textbf{CLIP}} & \multicolumn{2}{c}{\textbf{RemoteCLIP}} & \multicolumn{2}{c}{\textbf{GeoTreeCLIP}} \\
\cmidrule(lr){2-3} \cmidrule(lr){4-5} \cmidrule(lr){6-7}
 & Top-1 & Top-5 & Top-1 & Top-5 & Top-1 & Top-5 \\
\midrule
Africa ($n=558$)        & 17.20 $\pm$ 0.05 & 19.12 $\pm$ 0.05 & 0.00 $\pm$ 0.00 & 0.00 $\pm$ 0.00 & \textbf{17.31 $\pm$ 0.21} & \textbf{56.63 $\pm$ 1.51} \\
Asia ($n=1120$)         & 0.00 $\pm$ 0.00 & 0.00 $\pm$ 0.00 & 0.00 $\pm$ 0.00 & 4.91 $\pm$ 0.09 & \textbf{8.31 $\pm$ 0.19} & \textbf{39.61 $\pm$ 1.60} \\
Europe ($n=1648$)       & 0.00 $\pm$ 0.00 & 19.08 $\pm$ 0.00 & 0.00 $\pm$ 0.00 & 0.00 $\pm$ 0.00 & \textbf{19.96 $\pm$ 0.19} & \textbf{49.91 $\pm$ 0.24} \\
North America ($n=4442$)& 0.00 $\pm$ 0.00 & 1.15 $\pm$ 0.02 & 0.00 $\pm$ 0.00 & 6.81 $\pm$ 0.01 & \textbf{18.98 $\pm$ 0.33} & \textbf{52.79 $\pm$ 0.45} \\
Oceania ($n=812$)       & 0.92 $\pm$ 0.06 & 4.83 $\pm$ 0.07 & \textbf{13.25 $\pm$ 0.00} & 24.64 $\pm$ 0.18 & 7.64 $\pm$ 0.11 & \textbf{24.83 $\pm$ 0.02} \\
South America ($n=1150$)& 0.00 $\pm$ 0.00 & 14.76 $\pm$ 0.01 & 0.00 $\pm$ 0.00 & 4.60 $\pm$ 0.25 & \textbf{16.83 $\pm$ 0.75} & \textbf{41.89 $\pm$ 1.19} \\
\bottomrule
\end{tabular}%
}
\end{table}

GeoTreeCLIP demonstrates relatively consistent zero-shot capabilities across diverse geographic regions, with top-1 accuracies ranging from 7.64\% to 19.96\%. The highest performance is observed in Europe and North America, which aligns with the higher density of training samples available for these regions in the GlobalGeoTree dataset. Conversely, performance is lower in Asia and Oceania, likely reflecting the higher species diversity and relative scarcity of labeled data in these areas.

Notably, baseline models exhibit extreme regional bias. CLIP, for example, achieves high accuracy in Africa (17.20\%) but drops to 0.00\% in most other continents. This suggests that without domain-specific pretraining, models may rely on spurious correlations with broad landscape features (e.g., savannas) rather than learning discriminative species-level traits. GeoTreeCLIP's ability to maintain performance across all continents highlights its generalization capabilities.

\color{black}

\subsubsection{Few-shot evaluation}
\begin{table}[h]
\centering
\caption{Few-shot evaluation on \textit{GlobalGeoTree-10kEval}. Results are presented as mean accuracy (\%) $\pm$ standard deviation (\%) over 5 runs.}
\label{tab:few-shot}
\begin{tabular}{lcccccc}
\toprule
\multirow{2}{*}{\textbf{Taxon.}} & \multicolumn{2}{c}{\textbf{CLIP}} & \multicolumn{2}{c}{\textbf{RemoteCLIP}} & \multicolumn{2}{c}{\textbf{GeoTreeCLIP}} \\
\cmidrule(lr){2-3} \cmidrule(lr){4-5} \cmidrule(lr){6-7}
& Top-1 & Top-5 & Top-1 & Top-5 & Top-1 & Top-5 \\
\midrule
\multicolumn{7}{l}{\textit{One-Shot Evaluation}} \\
\midrule
Family & 2.95 $\pm$ 0.01 & 15.06 $\pm$ 0.03 & 11.25 $\pm$ 0.01 & 23.85 $\pm$ 0.03 & \textbf{29.37 $\pm$ 0.07} & \textbf{69.38 $\pm$ 0.34} \\
Genus  & 2.43 $\pm$ 0.01 &  8.14 $\pm$ 0.02 &  2.31 $\pm$ 0.01 &  7.68 $\pm$ 0.02 & \textbf{27.70 $\pm$ 0.11} & \textbf{64.40 $\pm$ 0.29} \\
Species& 1.67 $\pm$ 0.01 &  6.59 $\pm$ 0.03 &  1.94 $\pm$ 0.00 &  6.59 $\pm$ 0.02 & \textbf{25.80 $\pm$ 0.15} & \textbf{62.43 $\pm$ 0.25} \\
\midrule
\multicolumn{7}{l}{\textit{Three-Shot Evaluation}} \\
\midrule
Family & 6.19 $\pm$ 0.01 & 23.50 $\pm$ 0.03 & 4.44 $\pm$ 0.02 & 17.02 $\pm$ 0.04 & \textbf{37.77 $\pm$ 0.23} & \textbf{75.49 $\pm$ 0.25} \\
Genus  & 4.04 $\pm$ 0.01 & 12.76 $\pm$ 0.03 & 2.77 $\pm$ 0.03 & 11.70 $\pm$ 0.02 & \textbf{36.19 $\pm$ 0.22} & \textbf{72.50 $\pm$ 0.23} \\
Species& 3.41 $\pm$ 0.01 & 11.46 $\pm$ 0.03 & 1.88 $\pm$ 0.03 &  9.40 $\pm$ 0.06 & \textbf{33.67 $\pm$ 0.24} & \textbf{71.53 $\pm$ 0.23} \\
\bottomrule
\end{tabular}%
\end{table}

The results in Table \ref{tab:few-shot} demonstrate that providing even a small amount of labeled data for fine-tuning generally improves performance compared to the zero-shot setting across all models and taxonomic levels. GeoTreeCLIP consistently achieves the highest accuracies in both one-shot and three-shot scenarios. For instance, in the one-shot setting at the species level, GeoTreeCLIP reaches a top-1 accuracy of 25.80\%, substantially outperforming CLIP (1.67\%) and RemoteCLIP (1.94\%). This advantage becomes even more pronounced with three shots, where GeoTreeCLIP's species-level top-1 accuracy increases to 33.67\%, while CLIP and RemoteCLIP show more modest gains to 3.41\% and 1.88\%, respectively.

Our few-shot experiments reveal distinct patterns across models when increasing from one to three shots. GeoTreeCLIP demonstrates substantial improvements across all taxonomic levels (species top-1 accuracy rising from 25.80\% to 33.67\%), while CLIP shows consistent but smaller gains. RemoteCLIP exhibits mixed results, including a slight decrease in family-level accuracy, suggesting difficulties effectively utilizing additional examples. Both baseline models demonstrate limited capacity to leverage few-shot supervision compared to GeoTreeCLIP, with only marginal improvements over their zero-shot performances (Table \ref{tab:zero-shot}), particularly at finer taxonomic levels. This indicates that general pretraining approaches may not align sufficiently with the specific challenges of fine-grained tree species classification from remote sensing data, even when provided with in-domain examples.

The poor performance of CLIP and RemoteCLIP can likely be attributed to their design, which is optimized for RGB three-channel data and lacks the capability to process time-series information. Additionally, these models struggle with the small-patch classification tasks required for tree species identification. These limitations further emphasize the importance of introducing this benchmark for the global tree species classification task. More benchmark results on larger evaluation subsets can be found in Table \ref{tab:eval300_combined} and Table \ref{tab:eval900_combined} in Appendix \ref{performance_eval}.

\section{In-depth Analysis}
\subsection{Comparison with Supervised Learning Paradigm}
\label{app:supervised_comparison}

To further contextualize the performance of our contrastive learning-based GeoTreeCLIP model, we conducted an additional experiment using a traditional supervised learning paradigm. This allows for a more direct comparison of learning objectives (contrastive vs. supervised) while keeping core architectural components and training settings as consistent as possible.

\subsubsection{Supervised Model Architecture and Training}
\label{app:supervised_model_arch}

We designed a supervised model, termed SupervisedGeoTree, which retains the visual processing pathway of GeoTreeCLIP, including the \texttt{VisualEncoder} for Sentinel-2 time series and the \texttt{AuxiliaryEncoder} for environmental variables. The features from these two encoders are projected, normalized, and then fused via concatenation followed by a fusion layer, similar to the visual feature preparation in GeoTreeCLIP. 

However, unlike GeoTreeCLIP, the SupervisedGeoTree model does not include a text encoder or employ a contrastive loss. Instead, the fused visual-auxiliary features are fed into four independent classification heads (fully connected layers), each dedicated to predicting labels for one of the hierarchical taxonomic levels: functional type (\texttt{level0}), family (\texttt{level1\_family}), genus (\texttt{level2\_genus}), and species (\texttt{level3\_species}). The number of output neurons for each head corresponds to the number of unique classes at that respective taxonomic level in the GlobalGeoTree (4 for \texttt{level0}, 275 for family, 2,734 for genus, and 21,001 for species).

The model was trained on the \textit{GlobalGeoTree-6M} dataset. The loss function employed was a sum of standard Cross-Entropy losses, calculated independently for each of the four taxonomic levels. The contributions of each level's loss to the total loss were equally weighted. All other training hyperparameters, including the learning rate ($1 \times 10^{-5}$), optimizer (AdamW), weight decay, number of epochs (25), warmup strategy (5 epochs), learning rate scheduler (Cosine Annealing with Warm Restarts), and batch size, were kept identical to those used for pretraining GeoTreeCLIP to ensure a fair comparison of the learning paradigms.

\subsubsection{Comparison of the Supervised Model}
\label{app:supervised_results}
After training on the \textit{GlobalGeoTree-6M} dataset, the SupervisedGeoTree model was evaluated on the \textit{GlobalGeoTree-10kEval} subset. Since this model is trained with fixed classification heads for the classes seen during training, its ability to perform "zero-shot" classification in the same sense as a CLIP-style model (i.e., classifying entirely new, unseen samples provided at test time) is inherently limited. However, for this comparison, we evaluate its performance on the classes within \textit{GlobalGeoTree-10kEval} that were also part of the \textit{GlobalGeoTree-6M} training vocabulary for each respective taxonomic head. If a class in \textit{GlobalGeoTree-10kEval} was not in the training vocabulary for a specific head, it cannot be correctly predicted by that head.

Notably, the \textit{GlobalGeoTree-6M} dataset was designed to retain nearly all tree species categories, as our goal was to pretrain a model capable of classifying the full spectrum of tree species. Consequently, the zero-shot evaluation here can be seen as measuring the model's “zero-shot” transfer capability on unseen datasets, aligning with the concept of "in-domain" zero-shot classification defined in \cite{wang2024skyscript}.

\begin{table}[h]
\centering
\caption{Evaluation on \textit{GlobalGeoTree-10kEval}. The table compares our vision-language model, GeoTreeCLIP, against the CLIP and a traditional supervised baseline (SupervisedGeoTree). Results are mean accuracy (\%) $\pm$ standard deviation (\%) over 5 runs.}
\label{tab:zero-shot-sup}
\begin{tabular}{lcccccc} 
\toprule
\multirow{2}{*}{\textbf{Taxon.}} & \multicolumn{2}{c}{\textbf{CLIP}} & \multicolumn{2}{c}{\textbf{SupervisedGeoTree}} & \multicolumn{2}{c}{\textbf{GeoTreeCLIP}} \\
\cmidrule(lr){2-3} \cmidrule(lr){4-5} \cmidrule(lr){6-7} 
& Top-1 & Top-5 & Top-1 & Top-5 & Top-1 & Top-5 \\
\midrule
Family & 10.80 $\pm$ 0.03 & 25.32 $\pm$ 0.05 & 9.55 $\pm$ 0.10 & 27.41 $\pm$ 0.07 & \textbf{20.99 $\pm$ 0.28} & \textbf{56.88 $\pm$ 0.42} \\
Genus  &  1.09 $\pm$ 0.01 &  9.34 $\pm$ 0.01 & 1.19 $\pm$ 0.09 &  8.18 $\pm$ 0.14 & \textbf{18.39 $\pm$ 0.26} & \textbf{50.98 $\pm$ 0.41} \\
Species&  1.09 $\pm$ 0.01 &  7.02 $\pm$ 0.02 & 0.00 $\pm$ 0.00 &  0.28 $\pm$ 0.02 & \textbf{16.71 $\pm$ 0.25} & \textbf{47.52 $\pm$ 0.37} \\
\bottomrule
\end{tabular}%
\end{table}

The results in Table \ref{tab:zero-shot-sup} indicate that a consistent trend across all models is the decline in classification accuracy as the taxonomic level becomes finer (from family to species). This reflects the inherent increase in difficulty when distinguishing between more closely related taxa. However, the extent of this performance degradation varies significantly between models.

The SupervisedGeoTree model, which employs traditional supervised classification heads for each taxonomic level, achieves reasonable accuracy at the family level (9.55\% top-1). However, its performance drops sharply for genus (1.19\% top-1) and becomes negligible at the species level (0.00\% top-1). This drastic decline underscores the challenge of fine-grained classification when relying solely on visual and auxiliary features without leveraging the semantic relationships embedded in textual taxonomic labels, and the inherent limitation of generalizing to a large number of specific classes in a purely supervised manner.

In contrast, the proposed GeoTreeCLIP model demonstrates substantial improvements over all other baselines across every taxonomic level. At the family level, GeoTreeCLIP achieves a top-1 accuracy of 20.99\%, more than doubling the performance of CLIP and significantly outperforming SupervisedGeoTree. This advantage is even more pronounced at the finer-grained levels: GeoTreeCLIP obtains 18.39\% top-1 accuracy for genus and 16.71\% for species identification. These results strongly indicate the power of contrastive vision-language learning to align visual features with rich, hierarchical taxonomic text labels. This capability allows GeoTreeCLIP to learn more nuanced and generalizable representations, leading to its superior performance in this challenging zero-shot evaluation. The significant gap, especially compared to SupervisedGeoTree at the species level, highlights the efficacy of the contrastive learning paradigm for handling large, structured label spaces.

\subsection{Qualitative Analysis: t-SNE Feature Embeddings Visualization}

\label{app:tsne_visualization}
\begin{figure}[h]
    \centering
    \includegraphics[width=\textwidth]{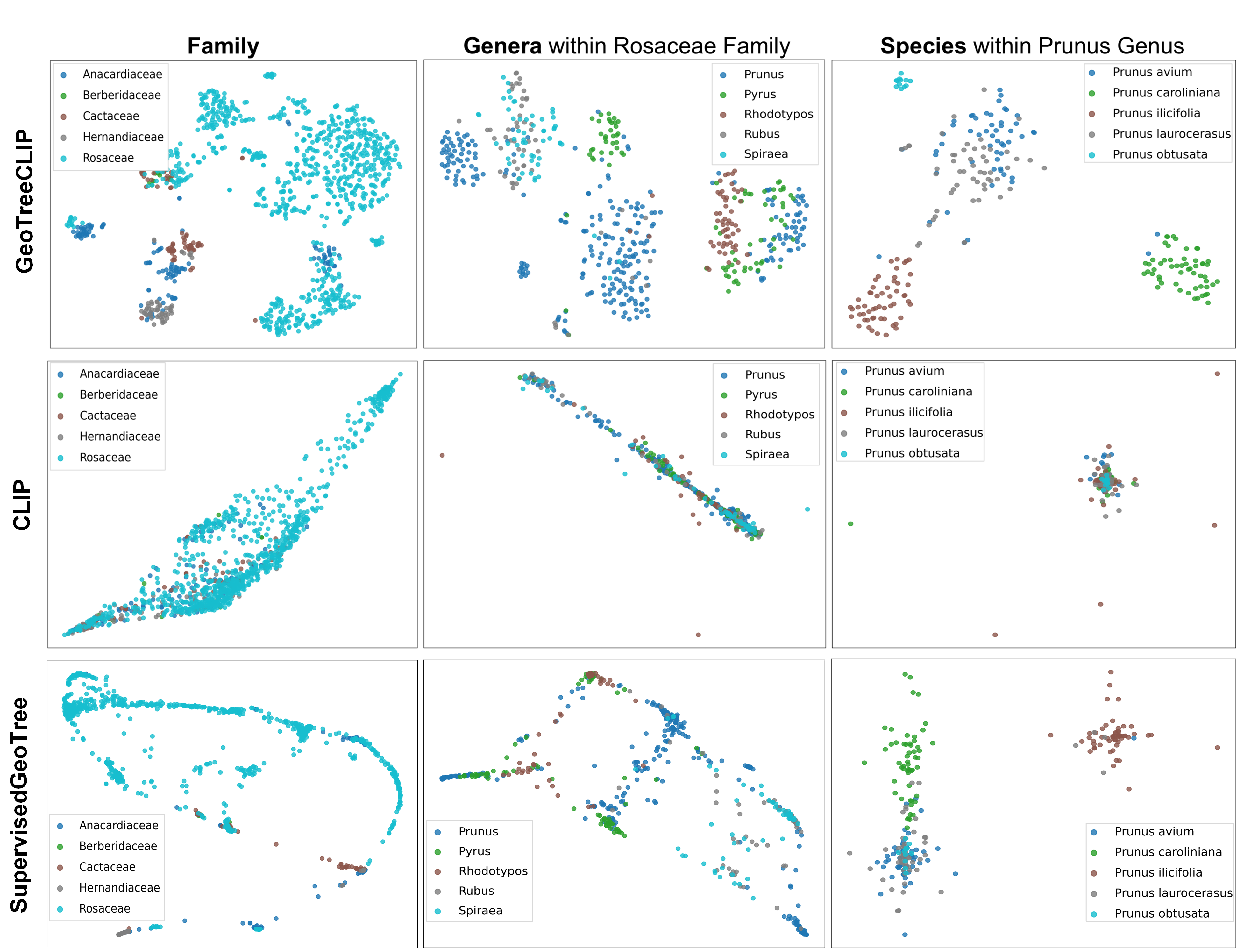} 
    \caption{t-SNE visualization of feature embeddings from \textit{GlobalGeoTree-10kEval-300} at different taxonomic levels, comparing GeoTreeCLIP (top row), CLIP (middle row), and SupervisedGeoTree (bottom row). Columns from left to right represent visualizations at the Family level (selected: Anacardiaceae, Berberidaceae, Cactaceae, Hernandiaceae, Rosaceae), Genus level (selected within Rosaceae: \textit{Prunus}, \textit{Pyrus}, \textit{Rhodotypos}, \textit{Rubus}, \textit{Spiraea}), and Species level (selected within \textit{Prunus}: \textit{Prunus avium}, \textit{Prunus caroliniana}, \textit{Prunus ilicifolia}, \textit{Prunus laurocerasus}, \textit{Prunus obtusata}).}
    \label{fig:tsne_all_models_comparison}
\end{figure}

To qualitatively assess and compare the learned feature representations from different modeling paradigms, we performed t-SNE visualizations \citep{van2008visualizing}. We used zero-shot image embeddings (or the final fused visual-auxiliary features for SupervisedGeoTree before the classification heads) extracted from a subset of the \textit{GlobalGeoTree-10kEval-300} dataset. This analysis includes our proposed GeoTreeCLIP, the CLIP pretrained by OpenAI, and the SupervisedGeoTree model. Given the large number of classes and the hierarchical nature of the labels, we adopted a selective visualization strategy: first examining embeddings at the family level for five randomly selected families (Anacardiaceae, Berberidaceae, Cactaceae, Hernandiaceae, and Rosaceae); then focusing on five genera within the Rosaceae family (\textit{Prunus}, \textit{Pyrus}, \textit{Rhodotypos}, \textit{Rubus}, and \textit{Spiraea}); and finally, visualizing five species within the \textit{Prunus} genus (\textit{Prunus avium}, \textit{Prunus caroliniana}, \textit{Prunus ilicifolia}, \textit{Prunus laurocerasus}, and \textit{Prunus obtusata}).

The comparative t-SNE visualizations are presented in Figure \ref{fig:tsne_all_models_comparison}. Across all three taxonomic levels (Family, Genus, and Species, shown as columns), GeoTreeCLIP (top row) consistently demonstrates the most effective separation and formation of distinct clusters. At the family level (left column), GeoTreeCLIP clearly distinguishes between the selected families. The CLIP (middle row) exhibits considerable overlap, particularly for the diverse Rosaceae family. SupervisedGeoTree (bottom row) shows some separation but less defined clusters compared to GeoTreeCLIP, with Rosaceae still forming a very broad distribution.

This pattern of superior clustering by GeoTreeCLIP continues at the genus level within Rosaceae (middle column). GeoTreeCLIP forms relatively distinct groups for genera like \textit{Prunus}, \textit{Pyrus}, and \textit{Rhodotypos}. Both CLIP and SupervisedGeoTree struggle more, with CLIP showing a highly condensed and overlapping structure, while SupervisedGeoTree offers some separation but with less clarity than GeoTreeCLIP. The most striking difference is observed at the species level within the \textit{Prunus} genus (right column). GeoTreeCLIP achieves remarkable separation, forming visually distinct clusters for each of the five \textit{Prunus} species. In contrast, both CLIP and SupervisedGeoTree largely fail to differentiate these closely related species, with their embeddings heavily intermingled.

These visualizations qualitatively affirm that GeoTreeCLIP, through its contrastive vision-language learning approach tailored with hierarchical taxonomic information and domain-specific data, learns more semantically meaningful and discriminative representations across all taxonomic ranks. This aligns with its superior quantitative performance in classification tasks compared to both general-domain VLMs and a traditional supervised approach.

\subsection{Ablation Study on Input Modalities}
To quantify the contribution of different input modalities, we conducted an extensive ablation study. We trained several variants of GeoTreeCLIP (for 10 epochs on \textit{GlobalGeoTree-6M}) and evaluated their zero-shot performance on the \textit{GlobalGeoTree-10kEval} benchmark. The results are summarized in Table~\ref{tab:ablation}.

\begin{table}[h]
\centering
\caption{Ablation study on input modalities. Performance is reported as mean Top-1 accuracy (\%) over 5 runs on \textit{GlobalGeoTree-10kEval}.}
\label{tab:ablation}
\begin{tabular}{lccc}
\toprule
\textbf{Model Variant} & \textbf{Family Acc.} & \textbf{Genus Acc.} & \textbf{Species Acc.} \\
\midrule
\textbf{GeoTreeCLIP (Full Input)} & \textbf{18.66\%} & \textbf{16.30\%} & \textbf{13.95\%} \\
- Image Only (10-band S2) & 15.83\% & 14.51\% & 12.62\% \\
- Auxiliary Only (27 vars) & 13.66\% & 10.50\% & 9.52\% \\
- RGB + Geolocation Only & 7.92\% & 6.04\% & 5.50\% \\
- RGB Only & 6.38\% & 5.08\% & 4.43\% \\
\bottomrule
\end{tabular}
\end{table}

The results lead to several key insights. First, the full input consistently outperforms all ablated versions, highlighting the synergistic effect of integrating spectral-temporal imagery with environmental context. Second, comparing the "Image Only (10-band)" and "RGB Only" variants reveals the critical importance of non-visible bands (e.g., NIR, SWIR) for vegetation analysis, as they provide a substantial performance boost (e.g., 15.83\% vs. 6.38\% at the family level). Third, adding geolocation to RGB imagery ("RGB + Geolocation") improves performance over "RGB Only", confirming that spatial priors help capture species' geographical ranges. The reason "Image Only (10-band)" outperforms "RGB + Geolocation" is that the rich information from the 7 additional spectral bands is more discriminative than the signal provided by geolocation alone.

\section{Ethics, Limitations and Impact}
\label{5}
\subsection{Ethics}
Our data collection process was designed to adhere to ethical standards for data reuse. All geolocated occurrence records sourced from GBIF \citep{lane2007global} were filtered to include only those with permissive licenses (e.g., CC0, CC-BY) explicitly chosen by the original data contributors. We also respected the privacy safeguards of the source platforms; for instance, the coordinates of sensitive or threatened species are often automatically obscured by platforms like iNaturalist, and our dataset preserves these privacy-preserving modifications.

The Sentinel-2 L2A data \citep{spoto2012overview} were accessed via Google Earth Engine. The auxiliary bioclimatic, geographic, and soil data were sourced from WorldClim \citep{fick2017worldclim}, USGS SRTM \citep{jarvis2008hole}, and SoilGrids \citep{poggio2021soilgrids}, respectively, all of which are publicly available.

\subsection{Limitations}
While GlobalGeoTree represents one of the largest and most comprehensive datasets of its kind, users should be aware of its inherent limitations. As with large part of datasets derived from citizen-science observations, it contains geographic and taxonomic biases. Data coverage is higher in regions with active observer communities (e.g., North America, Europe) and is skewed towards more common or easily identifiable species, resulting in a long-tail distribution. \textcolor{blue}{Furthermore, since we assimilated available official inventory data (e.g., NFIs) into the training set to maximize coverage, independent external validation remains challenging, and shared observation biases may influence evaluation metrics.} The ambiguous boundary between trees and shrubs in botanical classification further complicates the dataset, as some samples may represent shrubs rather than trees. Tree species taxonomy is also subject to frequent revisions driven by new genetic evidence, which may misalign dataset labels with updated classifications over time. Although CLIP-based models can identify unseen species to some extent, such taxonomic shifts may still affect model interpretability and evaluation consistency. Additionally, the dataset relies on Sentinel-2 data from 2020, restricting its ability to capture long-term vegetation dynamics or recent disturbances. Future versions incorporating multi-year observations could better account for phenological changes and climate-driven impacts.

\color{blue}
\subsection{Potential impact and Operational Deployment}
GlobalGeoTree holds significant potential for advancing forest monitoring, biodiversity conservation, and climate change mitigation. Beyond benchmarking, the dataset and GeoTreeCLIP model support practical operational workflows. For large-scale mapping, users can employ a sliding-window inference approach on Sentinel-2 tiles to generate pixel-wise species maps. To handle prediction uncertainty in diverse ecosystems, we recommend a hierarchical inference strategy, where the model defaults to genus or family-level classifications if species-level confidence falls below a threshold. Furthermore, to address regional domain shifts, the pretrained model serves as a robust foundation that can be efficiently fine-tuned with small sets of local field data. This "local adaptation" workflow allows practitioners to leverage global knowledge while tailoring predictions to specific forest management needs.
\color{black}

\section{Code and Data Availability}\label{dataaccess}
The GlobalGeoTree dataset is openly available under the Creative Commons Attribution 4.0 International (CC-BY 4.0) license. The complete dataset is archived on Hugging Face at \url{https://huggingface.co/datasets/yann111/GlobalGeoTree} under DOI \url{https://doi.org/10.15468/dd.9qxqyy} \citep{doi2025globalgeotree}. The pretraining and evaluation datasets are provided in WebDataset format \citep{aizman2019high}. This format enables efficient online data streaming to train models without requiring full dataset downloads, facilitating large-scale machine learning workflows. It also integrates seamlessly with popular deep learning frameworks, improving accessibility and usability for researchers.

All code is available in our github repository \url{https://github.com/MUYang99/GlobalGeoTree} under the Apache License 2.0, which provides comprehensive tools for using these resources. We are dedicated to maintaining and enhancing the dataset, addressing issues, and incorporating updates like new data sources or taxonomic revisions in future versions. The code and Huggingface repository will serve as the primary channels for updates and community feedback.

\section{Conclusion and future work}
In this paper, we introduced GlobalGeoTree, a large-scale, globally comprehensive dataset and benchmark for tree species classification. The dataset includes over 6 million geolocated tree occurrences spanning 21,001 species, paired with Sentinel-2 time series data and a rich set of auxiliary environmental variables. We also proposed GeoTreeCLIP, a baseline vision-language model specifically designed for this task, leveraging domain-specific pretraining on \textit{GlobalGeoTree-6M}. Experimental results demonstrate that GeoTreeCLIP significantly outperforms existing advanced models in classification accuracy across all taxonomic levels, highlighting both the effectiveness of our approach and the importance of introducing this benchmark for global tree species classification.

Future work could explore several promising directions. Expanding the GlobalGeoTree with more recent data, additional satellite sensors (e.g., SAR data for structural information), and a broader range of auxiliary variables could enhance its utility. \textcolor{blue}{Specifically, incorporating global structural datasets, such as GEDI LiDAR metrics or Meta's 1-m Canopy Height Map, could provide critical vertical information to distinguish between ecologically similar tree and shrub species, a limitation of current spectral-only approaches.} Investigating alternative vision-language model architectures, pretraining strategies, and methods for addressing the long-tail distribution could further improve classification accuracy, especially at the species level. Developing techniques for uncertainty estimation and improving model explainability are also critical areas for future work. Moreover, applying GlobalGeoTree and GeoTreeCLIP to real-world applications in biodiversity monitoring, conservation, and forest management could provide practical support and holds great potential.

\clearpage
\appendix
\section{GlobalGeoTree Dataset Statistics}
\label{app:ggt_stats}

The GlobalGeoTree dataset is a large-scale, multimodal resource for tree species classification. This section provides detailed statistics complementing the overview in the main paper.

\subsection{Data Records and Taxonomic Catalog}
To provide full transparency and facilitate detailed exploration of the dataset, we have made two key files publicly available in our Hugging Face repository (\url{https://huggingface.co/datasets/yann111/GlobalGeoTree/files}).

\begin{itemize}
    \item \textbf{Complete Taxonomic Catalog (\texttt{Tree\_species\_catalog.csv}):} This file contains the complete list of all 275 families, 2,734 genera, and 21,001 species included in GlobalGeoTree, along with their corresponding unique species keys used in the GBIF database. This catalog serves as a comprehensive reference for the taxonomic scope of our dataset.
    
    \item \textbf{Full Georeferenced Records (\texttt{GlobalGeoTree.csv}):} This file contains all 6.3 million geolocated records. Each row includes the coordinates, hierarchical taxonomic labels (from functional type to species), the original data source (e.g., iNaturalist Research-grade Observations), and other metadata as detailed in Table~\ref{tab:globalgeotree-features}. This allows for complete traceability and enables users to perform custom geographical or taxonomic subsetting.
\end{itemize}

\subsection{Basic Statistics}
The dataset encompasses a comprehensive collection of geolocated tree occurrences:
\begin{itemize}
    \item \textbf{Total Samples:} 6,263,345
    \item \textbf{Countries/Regions\footnote{The term "Countries/Regions" is used to align with global data standards (e.g., from GBIF) and accurately represents both sovereign states and other geographic entities such as overseas territories or dependencies, ensuring comprehensive coverage.} Covered:} 221
    \item \textbf{Taxonomic Coverage:} Families: 275, Genera: 2,734, Species: 21,001
\end{itemize}

\subsection{Long-tail Distribution Analysis}
\label{longtail}
\begin{figure}[h]
\centering
\includegraphics[width=0.99\textwidth]{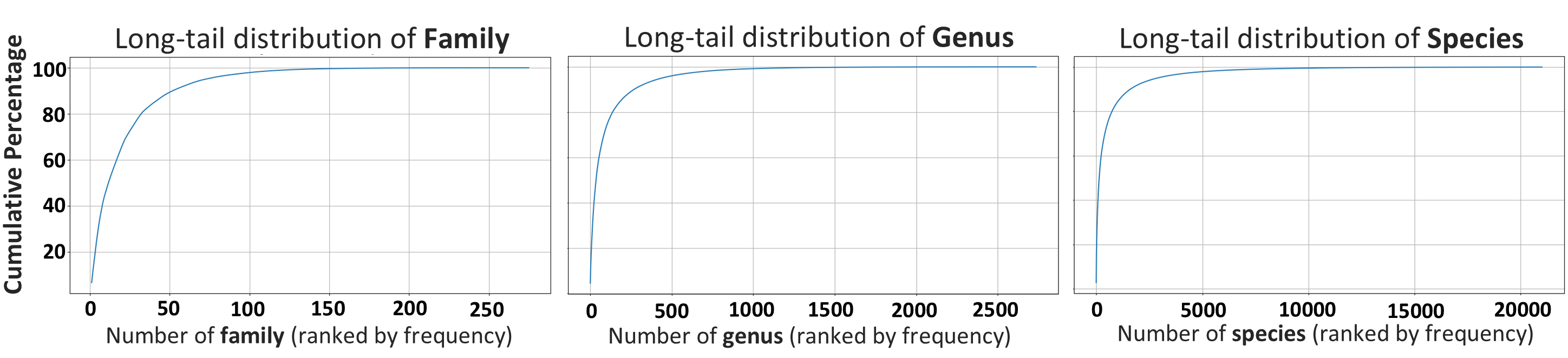}
\caption{Long-tail distribution across taxonomic levels in GlobalGeoTree.}
\label{fig:longtail}
\end{figure}

The dataset exhibits a characteristic long-tail distribution across taxonomic levels as shown in Fig. \ref{fig:longtail}. This highlights the challenge of classifying both common and rare taxa:
\begin{itemize}
    \item \textbf{Family Level:} The top 20\% of families (55 families) account for 91.01\% of all samples. Conversely, 24 families (8.73\% of total families) have fewer than 10 samples each.
    \item \textbf{Genus Level:} The top 20\% of genera (546 genera) cover 96.65\% of the samples. A significant portion, 760 genera (27.80\% of total genera), have fewer than 10 samples.
    \item \textbf{Species Level:} The distribution is most skewed at the species level, where the top 20\% of species (4,200 species) comprise 97.21\% of the samples. A majority of species, 11,611 species (55.29\% of total species), have fewer than 10 samples.
\end{itemize}
This long-tail distribution underscores the importance of evaluation strategies, like those employed for \textit{GlobalGeoTree-10kEval}, that explicitly consider species rarity.

\subsection{Detailed Categorical Statistics}
Below are statistics for key categorical attributes within the GlobalGeoTree dataset, illustrating geographical and taxonomic diversity and distribution.

\subsubsection{Location (\texttt{location})}
\begin{itemize}
    \item Number of unique countries/regions: 221
    \item Top 5 most frequent locations:
    \begin{enumerate}
        \item United States of America: 1,932,465
        \item Australia: 506,179
        \item Canada: 429,266
        \item Colombia: 330,896
        \item Russian Federation: 209,019
    \end{enumerate}
\end{itemize}

\begin{multicols}{2}
\subsubsection{Functional Type (\texttt{level0})}
\begin{itemize}
    \item Number of unique functional types: 4
    \item Distribution of functional types:
    \begin{itemize}
        \item Deciduous Broadleaf: 3,582,456
        \item Evergreen Broadleaf: 2,208,578
        \item Evergreen Needleleaf: 447,568
        \item Deciduous Needleleaf: 24,743
    \end{itemize}
\end{itemize}

\subsubsection{Taxonomic Family (\texttt{level1\_family})}
\begin{itemize}
    \item Number of unique families: 275
    \item Top 5 most frequent families:
    \begin{enumerate}
        \item Ericaceae: 423,365
        \item Fabaceae: 384,038
        \item Fagaceae: 362,317
        \item Rosaceae: 355,950
        \item Pinaceae: 320,415
    \end{enumerate}
\end{itemize}

\subsubsection{Taxonomic Genus (\texttt{level2\_genus})}
\begin{itemize}
    \item Number of unique genera: 2,734
    \item Top 5 most frequent genera:
    \begin{enumerate}
        \item \textit{Cornus}: 288,678
        \item \textit{Quercus}: 210,104
        \item \textit{Pinus}: 168,917
        \item \textit{Vaccinium}: 158,362
        \item \textit{Prunus}: 125,604
    \end{enumerate}
\end{itemize}

\subsubsection{Taxonomic Species (\texttt{level3\_species})}
\begin{itemize}
    \item Number of unique species: 21,001
    \item Top 5 most frequent species:
    \begin{enumerate}
        \item \textit{Cornus acuminata}: 180,120
        \item \textit{Securidaca volubilis}: 103,441
        \item \textit{Cupania sylvatica}: 99,797
        \item \textit{Bourreria cumanensis}: 96,304
        \item \textit{Fagus sylvatica}: 75,503
    \end{enumerate}
\end{itemize}
\end{multicols}

\color{blue}
\subsection{Representative Species Profile}
Table \ref{tab:rep_species} provides a structured summary of representative species from the GlobalGeoTree dataset, highlighting the diversity across different functional types and geographic regions.

\begin{table}[h]
\centering

\caption{\color{blue}Structured summary of representative tree species in GlobalGeoTree, spanning major functional types and biogeographical realms.}
\label{tab:rep_species}
\resizebox{\textwidth}{!}{%
\small
\begin{tabular}{llllll}
\toprule
\textbf{Functional Type} & \textbf{Family} & \textbf{Genus} & \textbf{Species} & \textbf{Dominant Region} & \textbf{Ecological Context} \\
\midrule
Evergreen Needleleaf & Pinaceae & \textit{Picea} & \textit{Picea abies} & Europe & Dominant boreal forest conifer \\
Evergreen Needleleaf & Pinaceae & \textit{Pseudotsuga} & \textit{Pseudotsuga menziesii} & North America & Key timber species of Pacific Northwest \\
Deciduous Broadleaf & Fagaceae & \textit{Quercus} & \textit{Quercus robur} & Europe & Keystone species of temperate woodlands \\
Deciduous Broadleaf & Sapindaceae & \textit{Acer} & \textit{Acer saccharum} & North America & Major component of northern hardwood forests \\
Evergreen Broadleaf & Myrtaceae & \textit{Eucalyptus} & \textit{Eucalyptus globulus} & Oceania & Fast-growing, widely distributed in Australia \\
Evergreen Broadleaf & Urticaceae & \textit{Cecropia} & \textit{Cecropia peltata} & South America & Pioneer species in neotropical rainforests \\
Deciduous Needleleaf & Pinaceae & \textit{Larix} & \textit{Larix decidua} & Europe & High-altitude conifer of the Alps \\
\bottomrule
\end{tabular}
}
\end{table}

\color{black}
\section{Auxiliary Environmental Variables}
\label{app:aux_env}
The 27 auxiliary variables used in GlobalGeoTree were sourced from three publicly available datasets, accessible via Google Earth Engine (GEE):

\begin{itemize}
    \item \textbf{WorldClim V1 (Bioclimatic Variables):} GEE Dataset ID: \texttt{WORLDCLIM/V1\_BIO}
    \item \textbf{SoilGrids 250m v2.0 (Soil Variables):} GEE Dataset ID: \texttt{ISRIC/soilgrids/v02}
    \item \textbf{USGS SRTM GL1 v003 (Geographic Variables):} GEE Dataset ID: \texttt{USGS/SRTMGL1\_003}
\end{itemize}

Table~\ref{tab:bioclim_vars} provides a detailed description of a subset of these variables.

\begin{longtable}{@{}lllp{7cm}@{}}
\caption{Description of the 19 bioclimatic variables from WorldClim (BIO1-BIO19) used in the GlobalGeoTree dataset.} \label{tab:bioclim_vars} \\

\toprule
\textbf{Variable} & \textbf{Units} & \textbf{Scale Factor} & \textbf{Description} \\
\midrule
\endfirsthead

\multicolumn{4}{c}%
{{\tablename\ \thetable{} -- continued from previous page}} \\
\toprule
\textbf{Variable} & \textbf{Units} & \textbf{Scale Factor} & \textbf{Description} \\
\midrule
\endhead

\midrule
\multicolumn{4}{r}{{Continued on next page}} \\
\midrule
\endfoot

\bottomrule
\endlastfoot

bio01 & \si{\degree}C & 0.1 & Annual Mean Temperature \\
bio02 & \si{\degree}C & 0.1 & Mean Diurnal Range (Mean of monthly (max temp - min temp)) \\
bio03 & \% & 1 & Isothermality (BIO2/BIO7 $\times$ 100) \\ 
bio04 & \si{\degree}C & 0.01 & Temperature Seasonality (Standard Deviation $\times$ 100) \\
bio05 & \si{\degree}C & 0.1 & Max Temperature of Warmest Month \\
bio06 & \si{\degree}C & 0.1 & Min Temperature of Coldest Month \\
bio07 & \si{\degree}C & 0.1 & Temperature Annual Range (BIO5 - BIO6) \\
bio08 & \si{\degree}C & 0.1 & Mean Temperature of Wettest Quarter \\
bio09 & \si{\degree}C & 0.1 & Mean Temperature of Driest Quarter \\
bio10 & \si{\degree}C & 0.1 & Mean Temperature of Warmest Quarter \\
bio11 & \si{\degree}C & 0.1 & Mean Temperature of Coldest Quarter \\
bio12 & mm & 1 & Annual Precipitation \\
bio13 & mm & 1 & Precipitation of Wettest Month \\
bio14 & mm & 1 & Precipitation of Driest Month \\
bio15 & Coeff. of Variation & 1 & Precipitation Seasonality \\
bio16 & mm & 1 & Precipitation of Wettest Quarter \\
bio17 & mm & 1 & Precipitation of Driest Quarter \\
bio18 & mm & 1 & Precipitation of Warmest Quarter \\
bio19 & mm & 1 & Precipitation of Coldest Quarter \\
\end{longtable}

\section{Details of Evaluation Subsets}
\label{eval_details}
\subsection{Overview and Construction}
\begin{figure}[H]
\centering
\subfloat[\textit{GlobalGeoTree-10kEval-300}]{
    \includegraphics[width=0.49\textwidth]{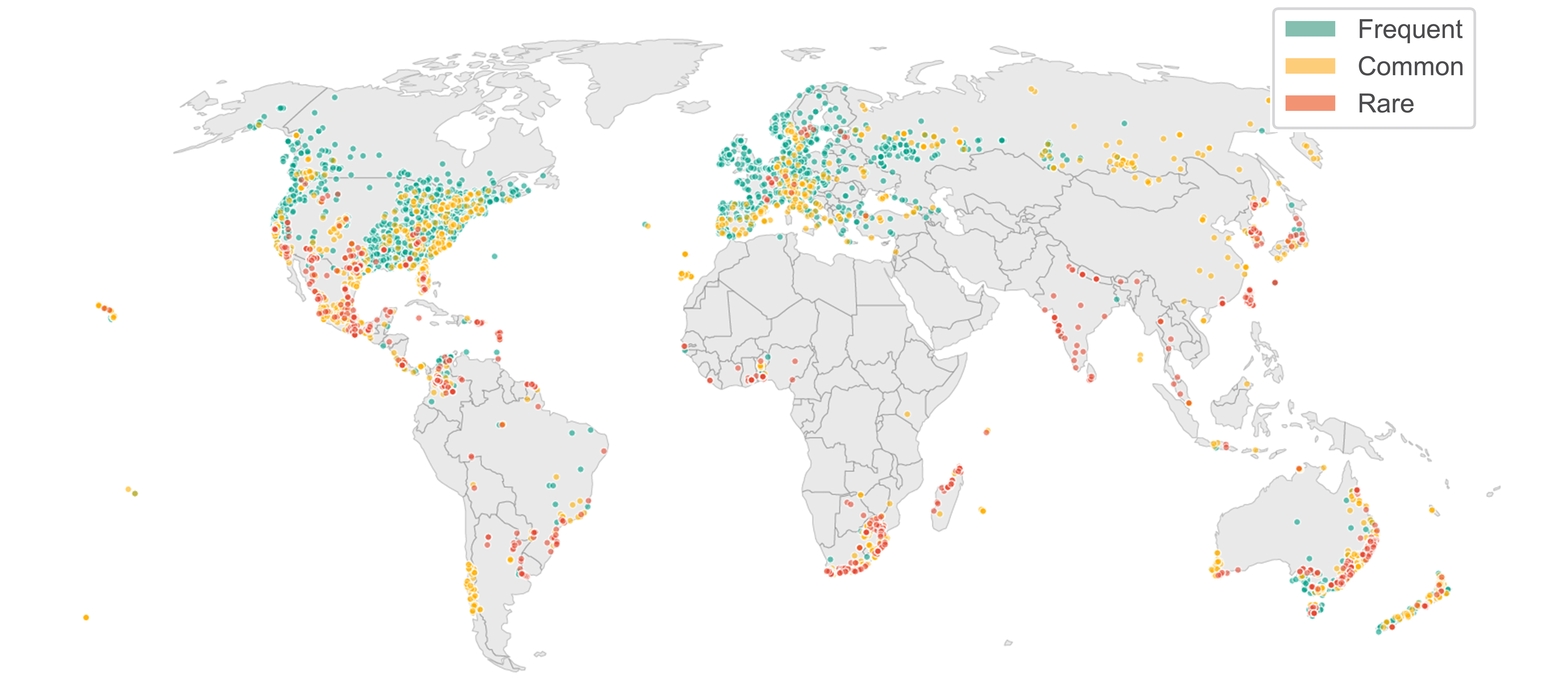}
    \label{fig:10kEval-300_distribution}
}
\hfill
\subfloat[\textit{GlobalGeoTree-10kEval-900}]{
    \includegraphics[width=0.49\textwidth]{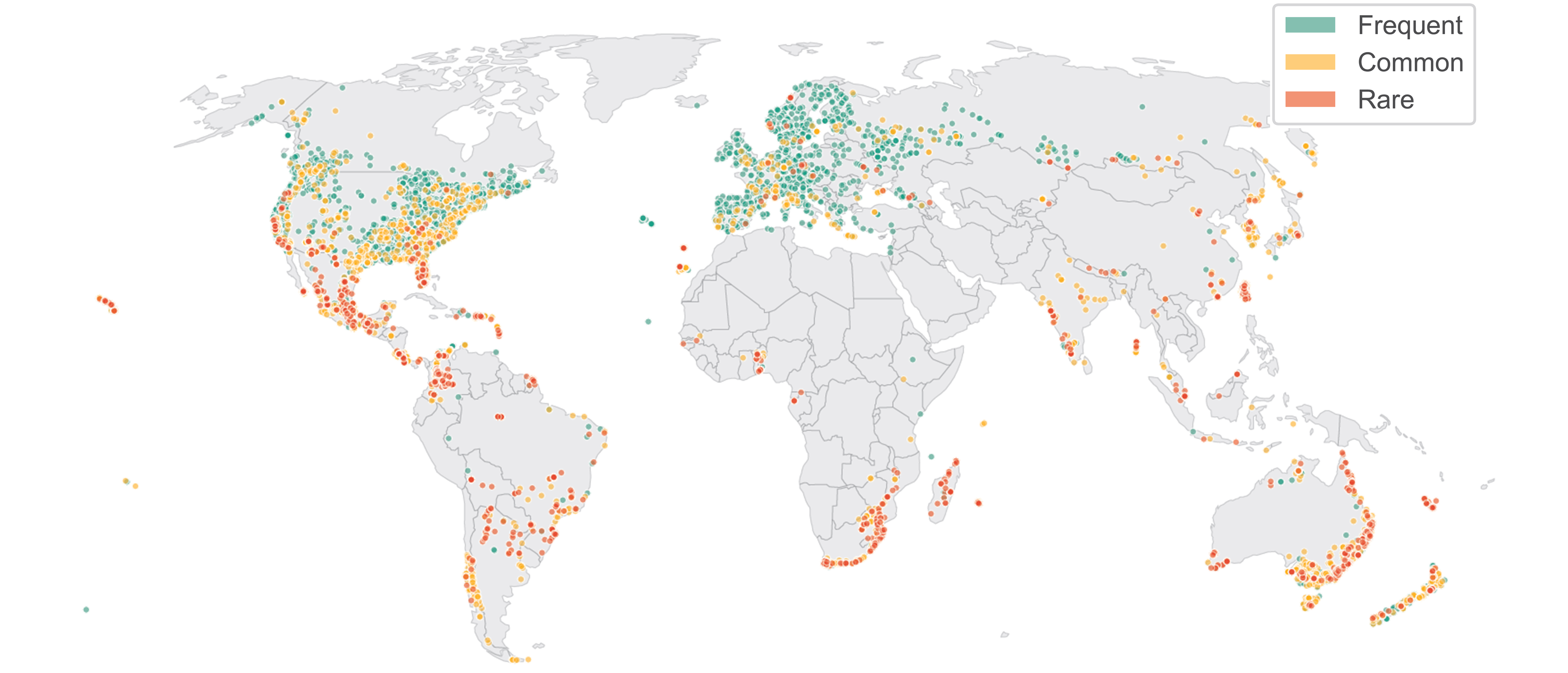}
    \label{fig:10kEval-900_distribution}
}
\caption{Geographic distributions of \textit{GlobalGeoTree-10kEval-300} and \textit{GlobalGeoTree-10kEval-900}.}
\label{fig:combined_distribution}
\end{figure}

To enable robust benchmarking across various taxonomic diversity scales and species rarity, we constructed three evaluation subsets: \textit{GlobalGeoTree-10kEval}, \textit{GlobalGeoTree-10kEval-300}, and \textit{GlobalGeoTree-10kEval-900}. These subsets were created by first categorizing all species in the GlobalGeoTree dataset into Rare, Common, and Frequent groups based on available sample counts (Sect. \ref{3.3}), then randomly selecting 30, 100, and 300 species per category, respectively. The primary \textit{GlobalGeoTree-10kEval} benchmark (90 species) is featured in the main paper, while the larger subsets enable assessment of model scalability and performance on increasingly complex tasks. Detailed overviews of each subset's composition are provided in Table \ref{tab:ggt_10keval_overview}, Table \ref{tab:ggt_10keval_300_overview}, and Table \ref{tab:ggt_10keval_900_overview}. The geographical distribution of the two additional evaluation sets is shown in Figure \ref{fig:10kEval-300_distribution} and Figure \ref{fig:10kEval-900_distribution}.

\subsection{\textit{GlobalGeoTree-10kEval-300} and \textit{GlobalGeoTree-10kEval-900}}

\begin{table}[ht]
\centering
\caption{Overview of the \textit{GlobalGeoTree-10kEval} evaluation subset. This subset comprises 30 species selected from each of the three rarity categories (Rare, Common, Frequent), totaling 90 unique species and 9,930 geolocated samples.}
\label{tab:ggt_10keval_overview}
\begin{tabular}{llr}
\toprule
\textbf{Category} & \textbf{Example Species} & \textbf{Num of Samples}  \\
\midrule
\textbf{Rare}     & \textit{Acacia platycarpa}        & 40                          \\
                  & \textit{Adenanthos cuneatus}      & 40                          \\
                  & \textit{Adenocarpus decorticans}  & 40                          \\
                  & \multicolumn{1}{c}{...}  & ...                         \\
\cmidrule(lr){1-3}
\textbf{Rare (Total)} & \textit{30 species selected} &   \textbf{1,200}            \\
\midrule
\textbf{Common}   & \textit{Abies religiosa}          & 110                             \\
                  & \textit{Aloe marlothii}           & 110                        \\
                  & \textit{Alternanthera sessilis}   & 110                             \\
                  & \multicolumn{1}{c}{...}  & ...                            \\
\cmidrule(lr){1-3}
\textbf{Common (Total)} & \textit{30 species selected} &  \textbf{3,300}            \\
\midrule
\textbf{Frequent} & \textit{Acer glabrum}             & 181                             \\
                  & \textit{Arctostaphylos glandulosa} & 181                          \\
                  & \textit{Ardisia paniculata}       & 181                           \\
                  & \multicolumn{1}{c}{...}  & ...                           \\
\cmidrule(lr){1-3}
\textbf{Frequent (Total)}& \textit{30 species selected} &    \textbf{5,430}            \\
\midrule
\textbf{Total}    & \textit{90 species total}  &    \textbf{9,930}           \\
\bottomrule
\end{tabular}
\end{table}

\begin{table}[H]
\centering
\caption{Overview of the \textit{GlobalGeoTree-10kEval-300} evaluation subset. This subset comprises 100 species selected from each of the three rarity categories (Rare, Common, Frequent), totaling 300 unique species and 10,000 geolocated samples.}
\label{tab:ggt_10keval_300_overview}
\begin{tabular}{llr}
\toprule
\textbf{Category} & \textbf{Example Species} & \textbf{Num of Samples}  \\
\midrule
\textbf{Rare}     & \textit{Abutilon wrightii}        & 12                          \\
                  & \textit{Acacia georgensis}        & 12                          \\
                  & \textit{Acacia loroloba}          & 12                          \\
                  & \multicolumn{1}{c}{...}  & ...                         \\
\cmidrule(lr){1-3}
\textbf{Rare (Total)} & \textit{100 species selected} &   \textbf{1,200}            \\
\midrule
\textbf{Common}   & \textit{Acacia confusa}           & 33                             \\
                  & \textit{Acacia mucronata}         & 33                        \\
                  & \textit{Achyranthes spec}         & 33                             \\
                  & \multicolumn{1}{c}{...}  & ...                            \\
\cmidrule(lr){1-3}
\textbf{Common (Total)} & \textit{100 species selected} &  \textbf{3,300}            \\
\midrule
\textbf{Frequent} & \textit{Acacia dealbata}          & 55                             \\
                  & \textit{Acacia decurrens}         & 55                          \\
                  & \textit{Acacia polyphylla}        & 55                           \\
                  & \multicolumn{1}{c}{...}  & ...                           \\
\cmidrule(lr){1-3}
\textbf{Frequent (Total)}& \textit{100 species selected} &    \textbf{5,500}            \\
\midrule
\textbf{Total}    & \textit{300 species total}  &    \textbf{10,000}           \\
\bottomrule
\end{tabular}
\end{table}

\begin{table}[ht]
\centering
\caption{Overview of the \textit{GlobalGeoTree-10kEval-900} evaluation subset. This subset comprises 300 species selected from each of the three rarity categories (Rare, Common, Frequent), totaling 900 unique species and 10,200 geolocated samples.}
\label{tab:ggt_10keval_900_overview}
\begin{tabular}{llr}
\toprule
\textbf{Category} & \textbf{Example Species} & \textbf{Num of Samples}  \\
\midrule
\textbf{Rare}     & \textit{Abies hickelii}           & 4                          \\
                  & \textit{Abutilon auritum}         & 4                          \\
                  & \textit{Acacia adunca}            & 4                          \\
                  & \multicolumn{1}{c}{...}  & ...                         \\
\cmidrule(lr){1-3}
\textbf{Rare (Total)} & \textit{300 species selected} &   \textbf{1,200}            \\
\midrule
\textbf{Common}   & \textit{Abies fraseri}            & 11                             \\
                  & \textit{Acacia echinula}          & 11                        \\
                  & \textit{Acacia falcata}           & 11                             \\
                  & \multicolumn{1}{c}{...}  & ...                            \\
\cmidrule(lr){1-3}
\textbf{Common (Total)} & \textit{300 species selected} &  \textbf{3,300}            \\
\midrule
\textbf{Frequent} & \textit{Abies amabilis}           & 19                             \\
                  & \textit{Abies balsamea}           & 19                          \\
                  & \textit{Acacia dealbata}          & 19                           \\
                  & \multicolumn{1}{c}{...}  & ...                           \\
\cmidrule(lr){1-3}
\textbf{Frequent (Total)}& \textit{300 species selected} &    \textbf{5,700}            \\
\midrule
\textbf{Total}    & \textit{900 species total}  &    \textbf{10,200}           \\
\bottomrule
\end{tabular}
\end{table}

\FloatBarrier
\section{Model Performance on \textit{GlobalGeoTree-10kEval-300} and \textit{GlobalGeoTree-10kEval-900}}
\label{performance_eval}
The evaluation results on the larger \textit{GlobalGeoTree-10kEval-300} (Table \ref{tab:eval300_combined}) and \textit{GlobalGeoTree-10kEval-900} (Table \ref{tab:eval900_combined}) subsets align with trends from the primary benchmark \textit{GlobalGeoTree-10kEval} (Table \ref{tab:zero-shot} and \ref{tab:few-shot}). GeoTreeCLIP consistently outperforms CLIP and RemoteCLIP across all settings (zero-shot, one-shot, three-shot) and taxonomic levels (Family, Genus, Species). Despite lower absolute accuracies on these challenging subsets, especially \textit{GlobalGeoTree-10kEval-900}, GeoTreeCLIP maintains a significant performance edge, highlighting the advantages of its domain-specific pretraining and tailored architecture for multimodal tree species classification.

\begin{table}[h!]
\centering
\caption{Zero-shot and Few-shot evaluation on \textit{GlobalGeoTree-10kEval-300}. Results are presented as mean accuracy (\%) $\pm$ standard deviation (\%) over 5 runs.}
\label{tab:eval300_combined}
\begin{tabular}{lcccccc}
\toprule
\multirow{2}{*}{\textbf{Taxon.}} & \multicolumn{2}{c}{\textbf{CLIP}} & \multicolumn{2}{c}{\textbf{RemoteCLIP}} & \multicolumn{2}{c}{\textbf{GeoTreeCLIP}} \\
\cmidrule(lr){2-3} \cmidrule(lr){4-5} \cmidrule(lr){6-7}
& Top-1 & Top-5 & Top-1 & Top-5 & Top-1 & Top-5 \\
\midrule
\multicolumn{7}{l}{\textit{Zero-Shot Evaluation}} \\
\midrule
Family & 7.07 $\pm$ 0.02 & 20.14 $\pm$ 0.03 & 1.35 $\pm$ 0.01 & 7.28 $\pm$ 0.08 & \textbf{12.55 $\pm$ 0.20} & \textbf{40.25 $\pm$ 0.30} \\
Genus  & 2.34 $\pm$ 0.01 & 5.59 $\pm$ 0.03 & 0.57 $\pm$ 0.01 & 2.10 $\pm$ 0.05 & \textbf{9.26 $\pm$ 0.24} & \textbf{28.34 $\pm$ 0.25} \\
Species& 0.46 $\pm$ 0.00 & 2.06 $\pm$ 0.01 & 0.56 $\pm$ 0.01 & 1.70 $\pm$ 0.02 & \textbf{7.87 $\pm$ 0.20} & \textbf{25.20 $\pm$ 0.21} \\
\midrule
\multicolumn{7}{l}{\textit{One-Shot Evaluation}} \\
\midrule
Family & 1.55 $\pm$ 0.00 & 6.80 $\pm$ 0.02 & 2.43 $\pm$ 0.01 & 10.30 $\pm$ 0.09 & \textbf{18.58 $\pm$ 0.21} & \textbf{50.41 $\pm$ 0.21} \\
Genus  & 0.87 $\pm$ 0.01 & 3.59 $\pm$ 0.01 & 0.66 $\pm$ 0.01 & 3.91 $\pm$ 0.07 & \textbf{14.57 $\pm$ 0.26} & \textbf{41.92 $\pm$ 0.27} \\
Species& 0.63 $\pm$ 0.00 & 2.35 $\pm$ 0.01 & 0.61 $\pm$ 0.01 & 2.75 $\pm$ 0.03 & \textbf{13.31 $\pm$ 0.22} & \textbf{38.39 $\pm$ 0.26} \\
\midrule
\multicolumn{7}{l}{\textit{Three-Shot Evaluation}} \\
\midrule
Family & 5.28 $\pm$ 0.01 & 16.12 $\pm$ 0.04 & 3.79 $\pm$ 0.01 & 11.49 $\pm$ 0.05 & \textbf{23.91 $\pm$ 0.26} & \textbf{57.97 $\pm$ 0.29} \\
Genus  & 1.79 $\pm$ 0.01 & 6.02 $\pm$ 0.02 & 1.11 $\pm$ 0.02 & 3.60 $\pm$ 0.04 & \textbf{19.22 $\pm$ 0.28} & \textbf{50.35 $\pm$ 0.27} \\
Species& 1.33 $\pm$ 0.00 & 4.17 $\pm$ 0.01 & 0.74 $\pm$ 0.02 & 2.64 $\pm$ 0.02 & \textbf{17.90 $\pm$ 0.23} & \textbf{47.54 $\pm$ 0.29} \\
\bottomrule
\end{tabular}%
\end{table}

\begin{table}[h!]
\centering
\caption{Zero-shot and Few-shot evaluation on \textit{GlobalGeoTree-10kEval-900}. Results are presented as mean accuracy (\%) $\pm$ standard deviation (\%) over 5 runs.}
\label{tab:eval900_combined}
\begin{tabular}{lcccccc}
\toprule
\multirow{2}{*}{\textbf{Taxon.}} & \multicolumn{2}{c}{\textbf{CLIP}} & \multicolumn{2}{c}{\textbf{RemoteCLIP}} & \multicolumn{2}{c}{\textbf{GeoTreeCLIP}} \\
\cmidrule(lr){2-3} \cmidrule(lr){4-5} \cmidrule(lr){6-7}
& Top-1 & Top-5 & Top-1 & Top-5 & Top-1 & Top-5 \\
\midrule
\multicolumn{7}{l}{\textit{Zero-Shot Evaluation}} \\
\midrule
Family & 4.77 $\pm$ 0.04 & 13.17 $\pm$ 0.05 & 1.10 $\pm$ 0.00 & 10.08 $\pm$ 0.02 & \textbf{7.62 $\pm$ 0.13} & \textbf{27.25 $\pm$ 0.05} \\
Genus  & 0.69 $\pm$ 0.01 & 2.58 $\pm$ 0.02 & 0.04 $\pm$ 0.00 & 0.55 $\pm$ 0.02 & \textbf{4.36 $\pm$ 0.10} & \textbf{16.32 $\pm$ 0.11} \\
Species& 0.12 $\pm$ 0.00 & 0.59 $\pm$ 0.00 & 0.04 $\pm$ 0.00 & 0.51 $\pm$ 0.02 & \textbf{3.40 $\pm$ 0.09} & \textbf{11.83 $\pm$ 0.05} \\
\midrule
\multicolumn{7}{l}{\textit{One-Shot Evaluation}} \\
\midrule
Family & 3.44 $\pm$ 0.02 & 9.76 $\pm$ 0.04 & 1.89 $\pm$ 0.04 & 4.69 $\pm$ 0.06 & \textbf{14.16 $\pm$ 0.26} & \textbf{41.81 $\pm$ 0.31} \\
Genus  & 1.53 $\pm$ 0.01 & 2.82 $\pm$ 0.02 & 0.25 $\pm$ 0.01 & 1.21 $\pm$ 0.02 & \textbf{10.24 $\pm$ 0.21} & \textbf{31.45 $\pm$ 0.32} \\
Species& 0.45 $\pm$ 0.01 & 1.27 $\pm$ 0.01 & 0.22 $\pm$ 0.01 & 0.83 $\pm$ 0.01 & \textbf{8.18 $\pm$ 0.15} & \textbf{25.17 $\pm$ 0.28} \\
\midrule
\multicolumn{7}{l}{\textit{Three-Shot Evaluation}} \\
\midrule
Family & 4.39 $\pm$ 0.02 & 12.93 $\pm$ 0.04 & 5.43 $\pm$ 0.09 & 13.29 $\pm$ 0.02 & \textbf{16.00 $\pm$ 0.14} & \textbf{46.07 $\pm$ 0.34} \\
Genus  & 2.08 $\pm$ 0.01 & 5.25 $\pm$ 0.02 & 0.77 $\pm$ 0.02 & 4.22 $\pm$ 0.04 & \textbf{12.50 $\pm$ 0.21} & \textbf{37.42 $\pm$ 0.20} \\
Species& 1.37 $\pm$ 0.01 & 3.97 $\pm$ 0.01 & 0.34 $\pm$ 0.02 & 1.84 $\pm$ 0.06 & \textbf{10.23 $\pm$ 0.25} & \textbf{31.79 $\pm$ 0.17} \\
\bottomrule
\end{tabular}%
\end{table}

\FloatBarrier        

\section{Additional Baselines Evaluated on \textit{GlobalGeoTree-10kEval}}
\label{app:additional_zeroshot_baselines_eval10k} 
To further contextualize GeoTreeCLIP's performance, we extended our zero-shot evaluation on the \textit{GlobalGeoTree-10kEval} subset to include two additional publicly available vision-language models: SkyCLIP-50 \citep{wang2024skyscript} and CLIP-laion-RS, a CLIP model pretrained on the remote sensing subset of LAION-2B \citep{schuhmann2022laion}. These models were evaluated on the \textit{GlobalGeoTree-10kEval} subset under the same zero-shot protocol used for CLIP and RemoteCLIP (features extracted from individual monthly images, probabilities averaged). For ease of comparison, their performance alongside our GeoTreeCLIP is presented in Table \ref{tab:additional_baselines_geotreeclip_eval10k}.

\begin{table}[h!]
\centering
\caption{Zero-shot evaluation of SkyCLIP-50, CLIP-laion-RS, and GeoTreeCLIP on \textit{GlobalGeoTree-10kEval}. Results are presented as mean accuracy (\%) $\pm$ standard deviation (\%) over 5 runs.}
\label{tab:additional_baselines_geotreeclip_eval10k} 
\begin{tabular}{lcccccc}
\toprule
\multirow{2}{*}{\textbf{Taxon.}} & \multicolumn{2}{c}{\textbf{SkyCLIP-50}} & \multicolumn{2}{c}{\textbf{CLIP-laion-RS}} & \multicolumn{2}{c}{\textbf{GeoTreeCLIP}} \\
\cmidrule(lr){2-3} \cmidrule(lr){4-5} \cmidrule(lr){6-7}
& Top-1 & Top-5 & Top-1 & Top-5 & Top-1 & Top-5 \\
\midrule
Family & 2.33 $\pm$ 0.01 & 18.40 $\pm$ 0.08 & 1.15 $\pm$ 0.01 & 17.83 $\pm$ 0.16 & \textbf{20.99 $\pm$ 0.28} & \textbf{56.88 $\pm$ 0.42} \\
Genus  & 1.10 $\pm$ 0.03 &  6.48 $\pm$ 0.04 & 1.12 $\pm$ 0.01 &  7.33 $\pm$ 0.03 & \textbf{18.39 $\pm$ 0.26} & \textbf{50.98 $\pm$ 0.41} \\
Species& 1.10 $\pm$ 0.03 &  6.36 $\pm$ 0.04 & 1.12 $\pm$ 0.01 &  7.27 $\pm$ 0.02 & \textbf{16.71 $\pm$ 0.25} & \textbf{47.52 $\pm$ 0.37} \\
\bottomrule
\end{tabular}%
\end{table}

The results in Table \ref{tab:additional_baselines_geotreeclip_eval10k} show that both SkyCLIP-50 and CLIP-laion-RS, despite their pretraining on remote sensing imagery, achieve zero-shot accuracies on \textit{GlobalGeoTree-10kEval} that are substantially lower than our GeoTreeCLIP. For instance, at the species level, SkyCLIP-50 obtains a top-1 accuracy of 1.10\% and CLIP-LAION-RS achieves 1.12\%, in contrast to GeoTreeCLIP's 16.71\%. As indicated in the main text for similar baseline models (CLIP, RemoteCLIP), such performance can partly be attributed to their design, which is often optimized for RGB data and lacks effective handling of time-series information or small-patch classification crucial for tree species identification. These limitations further emphasize the value of the GlobalGeoTree benchmark and the effectiveness of our tailored GeoTreeCLIP approach in advancing global tree species classification research.

The comparison highlights that general remote sensing pretraining alone is insufficient for the nuanced task of global fine-grained tree species identification. The domain-specific dataset characteristics, multimodal input integration (including time-series and auxiliary data), and the tailored contrastive learning approach of GeoTreeCLIP appear critical for achieving strong performance on this challenging benchmark.

\noappendix       




\appendixfigures  

\appendixtables   


\authorcontribution{Conceptualization: Y.M., Z.X., S.M., X.Z.; methodology: Y.M., Z.X., Y.W., S.M., X.Z.; software: Y.M., Z.X., Y.W., X.Z.; results validation: Y.M., Z.X., Y.W., S.M., F.E., H.K., M.K., X.Z.; analysis: Y.M., Z.X., X.Z.; data collection: Y.M., Y.W.; writing, and original draft preparation: Y.M.; paper revision: Y.M., Z.X., Y.W., S.M., F.E., H.K., M.K., X.Z.; visualization: Y.M., Z.X.; supervision: S.M., X.Z.; funding acquisition: X.Z.; project administration: X.Z.; resources: X.Z.} 

\competinginterests{none} 

\disclaimer{none} 

\begin{acknowledgements}
The work is jointly supported by the German Federal Ministry for Economic Affairs and Climate Action in the framework of the ”national center of excellence ML4Earth” (grant number: 50EE2201C), by the European Union’s Horizon Europe research and innovations actions programme in the framework of the ”Multi-source and Multi-scale Earth observation and Novel Machine Learning Methods for Mineral Exploration and Mine Site Monitoring (MultiMiner)” (grant number: 101091374) and “ThinkingEarth—Copernicus Foundation Models for a Thinking Earth” (Grant Agreement No. 101130544), by the German Federal Ministry for the Environment, Nature Conservation, Nuclear Safety and Consumer Protection (BMUV) based on a resolution of the German Bundestag (grant number: 67KI32002B; Acronym: \textit{EKAPEx}), by the European Research Council (ERC) (grant agreement No. [ERC-2022-PoC-101101093], Acronym: \textit{EO4FoodSecurity}) by the Excellence Strategy of the Federal Government and the Länder through the TUM Innovation Network EarthCare and by Munich Center for Machine Learning.
\end{acknowledgements}




\bibliographystyle{copernicus}
\bibliography{references}

\end{document}